\documentclass[twoside]{article}
\usepackage[accepted]{aistats2026}
\usepackage{amsmath,amssymb,amsfonts}
\usepackage{algorithm}
\usepackage{algpseudocode}
\usepackage{graphicx}
\usepackage{textcomp}
\usepackage{xcolor}
\usepackage{amsthm}
\theoremstyle{plain}
\usepackage[switch]{lineno} 
\usepackage{multirow}
\usepackage{url}
\usepackage{booktabs}
\usepackage{float}
\usepackage{placeins}
\usepackage{paralist}
\newtheorem{theorem}{Theorem}

\usepackage[round]{natbib}

\begin{document}

%
\runningtitle{GMRF Multi-Component VAE}

%
\runningauthor{Oubari, El Baha, Meunier, D\'ecatoire, Mougeot}

\twocolumn[
\aistatstitle{Multi-Component VAE with Gaussian Markov Random Field}

\aistatsauthor{Fouad Oubari$^{1,2}$ \quad Mohamed El Baha$^{3}$ \quad Rapha\"el Meunier$^{2}$ \\ 
\textbf{Rodrigue D\'ecatoire}$^{2}$ \quad \textbf{Mathilde Mougeot}$^{1,3}$}

\aistatsaddress{$^{1}$Université Paris-Saclay, CNRS, ENS Paris-Saclay, Centre Borelli \enspace
$^{2}$Michelin \enspace
$^{3}$ENSIIE\\[1pt]
\small\texttt{oubarifouad@gmail.com}\enspace
\small\texttt{elbahamohamed@gmail.com}\enspace
\small\texttt{raphael.meunier@michelin.com}\\
\small\texttt{rodrigue.decatoire@michelin.com}\enspace
\small\texttt{mathilde.mougeot@ens-paris-saclay.fr}} ]

\begin{abstract}
Multi-component datasets with intricate dependencies challenge current generative modeling techniques. Existing Multi-component Variational AutoEncoders rely on simplified aggregation strategies that compromise structural coherence across generated components. We introduce the Gaussian Markov Random Field Multi-Component Variational AutoEncoder, embedding Gaussian Markov Random Fields into both prior and posterior distributions to explicitly model cross-component relationships, enabling richer representation and faithful reproduction of complex interactions. Empirically, our model achieves state-of-the-art performance on a synthetic Copula dataset designed for intricate component relationships, competitive results on PolyMNIST, and significantly enhanced structural coherence on the real-world BIKED dataset.\footnote{\textcolor{black}{Code available at \url{https://github.com/foubari/gmrf_mcvae}.}}
\end{abstract}

\section{INTRODUCTION}
\label{sec:intro}

\begin{figure*}[!t]
\centering
\includegraphics[width=.69\textwidth]{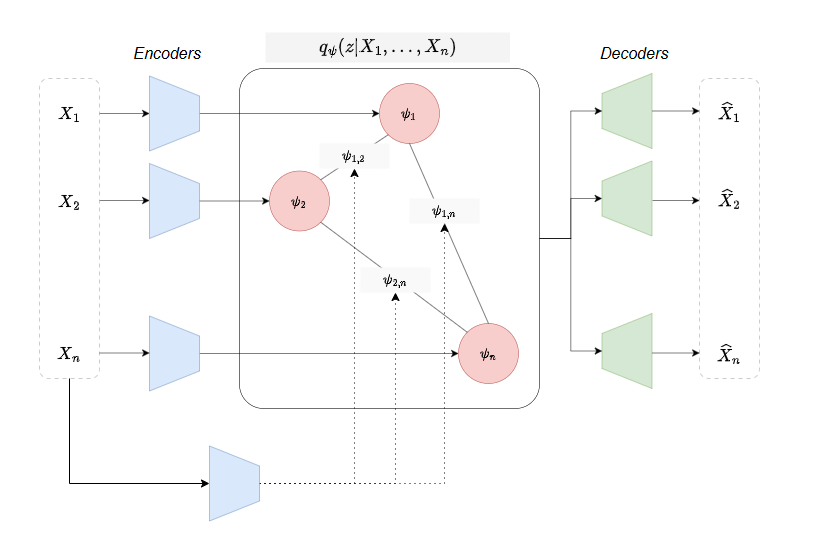}
\caption{A general MRF-based Multi-Component VAE: each component is assigned its own encoder-decoder pair, where the encoder learns unary potentials \(\psi_i\). A global encoder models pairwise potentials \(\psi_{i,j}\) among components. Sampling \(\mathbf{z}\) from this MRF-based latent space captures cross-component relationships. In practice, we adopt a Gaussian assumption for computational simplicity.}
\label{fig:gmrf_mvae}
\end{figure*}

\noindent
Multi-component datasets, ranging from medical imaging (CT-MRI pairs) \citep{puhr2019added,rahimi2022ct} to finance (multiple correlated markets) \citep{xie2024pixiu,lee2020multimodal} and industrial design (structured assemblies) \citep{cobb2023aircraftverse,10.1007/978-3-031-62281-6_17} are prevalent in real-world applications. Generating such data requires models that capture both component-wise intricacies and cross-component interactions. 

Although various generative approaches exist for multimodal data \citep{wu2018multimodal,shi2019variational}, many rely on simplified aggregation schemes that overlook nuanced inter-component factors. Hence, we investigate Markov Random Fields (MRFs) as a structured way to encode the relationships more explicitly. By decomposing the latent distribution into terms that reflect how components relate to each other \citep{koller2009probabilistic}, MRFs allow more nuanced interactions to be embedded, potentially leading to generations that better preserve global consistency. 
\textcolor{black}{This is particularly relevant in conditional multi-component generation, where observing one component should constrain the plausible configurations of the others, and where the coherence of the assembled system is a central requirement.}

\vspace{-1.5ex}
\subsection{Contributions}
\textcolor{black}{
We propose the Gaussian Markov Random Field Multi-Component VAE (GMRF MCVAE) with the following contributions:}

\begin{inparaenum}[(i)]
    \item \textcolor{black}{\textbf{Gaussian MRF Multi-Component VAE:} Novel architecture embedding GMRFs in both prior and posterior for richer inter-component correlations.}
    \item \textcolor{black}{\textbf{Practical Covariance Construction:} Blockwise assembly guaranteeing symmetric positive definiteness with closed-form conditional generation.}
    \item \textbf{Robust Empirical Validation:} State-of-the-art on synthetic Copula data, competitive results on PolyMNIST, and improved structural coherence on real-world BIKED, with competitive FID.
    
\end{inparaenum}

\section{RELATED WORK}
\label{sec:related}
\subsection{Multi-Component VAEs}

The field of multi-component generative models has seen substantial growth recently.
Within this domain, VAE-based models have distinguished themselves due to their rapid and tractable sampling capabilities \citep{vahdat2020nvae}, as well as their robust generalization performance \citep{mbacke2024statistical}. The essence of multi-component generation lies in its ability to learn a joint latent representation from multiple data components, encapsulating a unified distribution. Traditional MCVAE frameworks typically adopt a structure with separate encoder/decoder pairs for each component, coupled with an aggregation mechanism to encode a cohesive joint representation across all components. A variety of methodologies have been introduced to synthesize these distributions within the latent space.

A seminal approach by \citet{wu2018multimodal} suggests that the joint latent posterior can be effectively approximated through using the Product of Experts (PoE) assumption. This strategy facilitates the generation of cross components at inference time without requiring an additional inference network or a multistage training process, marking a significant advancement over previous methodologies \citep{suzuki2016joint,vedantam2017generative}. However, this approach implicitly relies on the assumption that the posterior distribution can be approximated by factorizing distributions. This assumption presupposes independence among components, which may not be true. This assumption overlooks the complex inter-component relationships intrinsic to the data, potentially limiting the model's ability to fully capture the richness of multi-component interactions. 

An alternative framework proposed by \citet{shi2019variational} employs a Mixture of Experts (MoE) strategy for aggregating marginal posteriors. This method stands in contrast to the approach used in the MVAE \citep{wu2018multimodal}, which, according to the authors, is susceptible to a 'veto phenomenon' a scenario where an exceedingly low marginal posterior density significantly diminishes the joint posterior density. In contrast, the MoE paradigm mitigates the risk associated with overly confident experts by adopting a voting mechanism among experts, thereby distributing its density among all contributing experts. However, a critique by \citet{palumbo2023mmvae} highlights a fundamental limitation of the MMVAE approach: it tends to average the contribution of each component. Given that the model employs each component-specific encoder to reconstruct all other components, the resultant encoding is biased towards information that is common across all components. This bias towards commonality potentially undermines the model's ability to capture and represent the diversity inherent in multi-component datasets.


The mix of experts' products (MoPoE) framework \citep{sutter2021generalized} refines and generalizes the aggregation strategies of PoE and MoE, combining the precise joint posterior approximation of PoE with the improved learning of component-specific posteriors by MoE. The MoPoE model is designed to enhance multi-component learning by integrating these traits. Despite its conceptual advancements, the MoPoE model introduces a computational challenge due to its training strategy. It necessitates the evaluation of all conceivable component subsets, which equates to $2^M-1$ training configurations for $M$ components. This comprehensive strategy, while beneficial for robust learning across varied component combinations, leads to an exponential increase in computational requirements relative to the number of components. This aspect marks a significant limitation, especially for applications involving a large number of components.


To mitigate the averaging problem observed in mixture-based models, several studies \citep{sutter2020multimodal,palumbo2023mmvae} have adopted component-specific latent spaces. Specifically, \citet{palumbo2023mmvae} identifies a 'shortcut' phenomenon, characterized by information predominantly circulating within component-specific subspaces. To address this, an enhancement of \citet{shi2019variational}'s  model incorporates component-specific latent spaces designed exclusively for self-reconstruction. This strategy prevents the 'shortcut' by using a shared latent space to aggregate and a component-specific space to reconstruct unobserved components,  ensuring that only joint information is retained in the shared space. Despite this advancement over prior approaches by resolving the shortcut dilemma, the outlined method introduces a training procedure that encompasses both reconstruction and cross-reconstruction tasks for each component pairing, leading to a computational requirement of $M^2$ forward passes for $M$ components.

\subsection{Markov Random Fields}

Undirected Graphical Models, also called Markov Random Fields \citep{wainwright2008graphical,koller2009probabilistic,murphy2012machine}, were introduced to probability theory as a way to extend Markov processes from a temporal framework to a spatial one; they represent a stochastic process that has its origins in statistical physics \citep{kindermann1980markov}. Graphical models, including MRF, are notoriously hard to train due to the intractability of the partition function. This has led to numerous studies \citep{carreira2005contrastive,vuffray2020efficient,bach2002learning,tan2014learning,welling2005learning} aimed at developing more efficient methods for learning graphical models, including MRF.

\subsection{MRF in Machine Learning}
\vspace{-0.9ex}
Markov Random Fields have predominantly been used in image processing tasks such as image deblurring \citep{perez1998markov}, completion, texture synthesis, and image inpainting \citep{komodakis2007image}, as well as segmentation \citep{krahenbuhl2011efficient,bello1994combined}. However,
recent advancements in more efficient methodologies have led to a decline in the use of MRF, due to the relative complexity involved in their learning processes. 

To the best of our knowledge, there are limited instances where Markov Random Fields have been integrated within generative neural networks. Among these, \citet{johnson2016composing} introduced the Structured Variational AutoEncoder (SVAE), which combines Conditional Random Fields with Variational AutoEncoders to address a variety of data modeling challenges. The SVAE has been applied to discrete mixture models, latent linear dynamical systems for video data, and latent switching linear dynamical systems for behavior analysis in video sequences. This approach employs mean field variational inference to approximate the Evidence Lower Bound, targeting specific data types without explicitly focusing on inter-component relationships.

Similarly, \citet{khoshaman2018gumbolt} integrates Boltzmann Machines as priors within VAEs, focusing on discrete variables to model complex and multi-component distributions. Their methodology suggests either factorial or hierarchical structures for the posterior distribution, aiming to effectively model complex and multi-component distributions.

Although significant advances have been made, the application of MRF within the domain of multi-component generative models, particularly in enhancing the integration and modeling of complex dependencies among multiple components remains largely unexplored. Our work seeks to bridge this gap by proposing a novel integration of MRF within a Multi-Component Variational AutoEncoder framework, aimed at capturing the intricate inter-component relationships more effectively. This approach not only leverages the strengths of MRF but also addresses the limitations observed in existing multi-component generative \mbox{models.}

\subsection{Diffusion-based Multimodal Generation}

\textcolor{black}{
More recently, diffusion-based methods have also been explored for multimodal generation, especially in audio--video settings. For instance, \citet{ruan2023mm} introduce a joint multimodal diffusion process for synchronized audio--video generation. \citet{xing2024seeing} connect pretrained unimodal generators through diffusion latent aligners to improve cross-modal generation and alignment. More recently, \citet{hayakawa2024mmdisco} propose a discriminator-guided cooperative diffusion framework for joint audio--video generation.}

\textcolor{black}{In this paper, however, we do not further investigate diffusion-based models and instead focus on the VAE family, where our goal is to improve multi-component coherence through explicit structured latent dependencies.}

\vspace{-3ex}

\section{METHODS}
\label{seq:method}

We define $\mathbf{X} = (\mathbf{x}_1, \dots, \mathbf{x}_M)$ as a collection of $M$ random variables, each representing a distinct component. Our approach employs a Multi-Component Variational AutoEncoder with an integrated Markov Random Field in its latent space, specifically designed to effectively capture the complex inter-component relationships.

\subsection{Variational AutoEncoders}
\textcolor{black}{Variational AutoEncoders (VAEs) \citep{kingma2013auto} use variational inference to approximate intractable posteriors \(p(\mathbf{z}| \mathbf{X})\) by maximizing the Evidence Lower Bound (ELBO):
\begin{small}
\begin{equation}
\label{eq:elbo}
\text{ELBO} \;=\;
\mathbb{E}_{q_{\phi}(\mathbf{z}\mid \mathbf{X})}
     \bigl[\ln p_{\theta}(\mathbf{X}\mid \mathbf{z})\bigr]
\;-\;
\text{KL}\!\Bigl(
   q_{\phi}(\mathbf{z}\mid \mathbf{X})
   \;\big\|\;
   p(\mathbf{z})
\Bigr)
\end{equation}
\end{small}}
where \(q_{\phi}(\mathbf{z}| \mathbf{X})\) is the encoder's variational posterior and \(p_{\theta}(\mathbf{X}\mid \mathbf{z})\) is the decoder's likelihood.
\textcolor{black}{In the multi-component setting we adopt a factorized likelihood $p_\theta(\mathbf{X}\mid \mathbf{z})=\prod_{i=1}^M p_{\theta_i}(\mathbf{x}_i\mid \mathbf{z}_i)$. When both $p(\mathbf{z})$ and $q_\phi(\mathbf{z}|\mathbf{X})$ are multivariate Gaussians ($\boldsymbol{\mu}_p,\boldsymbol{\Sigma}_p$ and $\boldsymbol{\mu}_q,\boldsymbol{\Sigma}_q$), the KL divergence admits the closed form}
\begin{small}
\begin{equation}
\label{eq:kl_gaussian}
\textcolor{black}{\mathrm{KL}(q\|p)
=\tfrac{1}{2}\bigl[\mathrm{tr}(\boldsymbol{\Sigma}_p^{-1}\boldsymbol{\Sigma}_q)
+ \boldsymbol{\delta}^\top\!\boldsymbol{\Sigma}_p^{-1}\boldsymbol{\delta}
- K + \ln\!\tfrac{|\boldsymbol{\Sigma}_p|}{|\boldsymbol{\Sigma}_q|}\bigr],}
\end{equation}
\end{small}
\textcolor{black}{with $\boldsymbol{\delta}=\boldsymbol{\mu}_p-\boldsymbol{\mu}_q$ and $K$ the total latent dimension. Sampling uses the Cholesky factor $\boldsymbol{\Sigma}=\mathbf{L}\mathbf{L}^\top$: $\mathbf{z}=\boldsymbol{\mu}+\mathbf{L}\boldsymbol{\epsilon}$, $\boldsymbol{\epsilon}\sim\mathcal{N}(\mathbf{0},\mathbf{I})$, so that gradients flow through $(\boldsymbol{\mu},\mathbf{L})$.}

\subsection{Markov Random Fields} 
MRFs offer an intuitive framework to model inter-component dependencies, with nodes representing random variables and edges capturing their relationships. 

Mathematically, an MRF is defined over an undirected graph $G = (V, E)$ where each node corresponds to a random variable in the set $(\mathbf{z}) = \{\mathbf{z}_i\}_{i=1}^n$. 
The joint distribution over these random variables is specified in terms of potential functions over cliques (fully connected subgraphs) of $G$. A general mathematical definition of an MRF is given by \citet{murphy2012machine}:
\begin{small}
\begin{equation}
\label{eq:mrf_general_expression}
    p_{\scriptscriptstyle MRF} = \frac{1}{\mathcal{Z}}\exp \left[ - \sum_{C \in \mathcal{C}} \psi_C(\mathbf{z}_C) \right]
\end{equation}
\end{small}

where $\mathcal{C}$ is the set of cliques in the graph, $\psi_C$ are the potential functions that map configurations of the random variables within the clique to a real number, $\mathbf{z}_C$ denotes the set of random variables in clique $C$, and $\mathcal{Z}$ is the partition function that normalizes the distribution. In the context of our work, we model both the prior $p(\mathbf{z})$ and the posterior $q_{\phi}(\mathbf{z}|\mathbf{x}_1, \dots, \mathbf{x}_M)$ as fully connected MRF represented by unary $\psi_i(\mathbf{z}_i)$ and pairwise $\psi_{i,j}(\mathbf{z}_i,\mathbf{z}_j)$ potentials. This leads to the specific form:
\begin{small}
\begin{equation}
\label{eq:mrf_specific}
    p_{\scriptscriptstyle MRF}(\mathbf{z}) = \frac{1}{\mathcal{Z}}\exp \left[ - \left(\sum_{i<j}^M \psi_{i,j}(\mathbf{z}_i,\mathbf{z}_j)+\sum_{i}^M \psi_{i}(\mathbf{z}_i)\right) \right]
\end{equation}
\end{small}

\noindent with $\mathbf{z} = (\mathbf{z}_1,\cdots,\mathbf{z}_M)$. This formulation enables the modeling of dependencies between components in our multi-component VAE framework by leveraging the structure of MRFs to capture both local and global interactions within the latent space.

\subsection{General MRF MCVAE Architecture and the Gaussian Assumption}
\label{seq:mrf_mvae}

\subsubsection{General Framework}  

In order to incorporate the MRF representation into a multi-component VAE framework, we employ the architecture illustrated in Figure \ref{fig:gmrf_mvae}.  Each component \(\mathbf{x}_i\) is paired with an encoder-decoder structure: the component-specific encoder outputs the parameters of the unary potentials \(\psi_i\), while a global encoder produces the pairwise potentials \(\psi_{i,j}\). Together, these potentials define the joint posterior \(q_\phi(\mathbf{z} | \mathbf{x}_1, \dots, \mathbf{x}_M)\), capturing both individual component details and their relationships. The latent variable \(\mathbf{z} = (\mathbf{z}_1, \dots, \mathbf{z}_M)\) is sampled from this posterior and decomposed into sub-vectors \(\mathbf{z}_i\), which are then reconstructed by their respective decoders.
\textcolor{black}{However, the MRF framework introduces two significant challenges: (i) computing the partition function \(\mathcal{Z}\) is computationally intractable in fully connected MRFs, and (ii) sampling from generalized MRFs is inherently complex, hindering gradient-based optimization. These limitations motivate our adoption of the Gaussian assumption described next.}



\subsubsection{GMRF MCVAE}
\textcolor{black}{These challenges lead us to adopt a Gaussian assumption for tractability. In standard GMRF notation \citep{murphy2012machine}, the unary and pairwise potentials become $\psi_i(\mathbf{z}_i) = \exp\!\bigl(-\tfrac12\,\mathbf{z}_i^\top\Lambda_{i,i}\,\mathbf{z}_i + \eta_i^\top \mathbf{z}_i\bigr)$ and $\psi_{i,j}(\mathbf{z}_i,\mathbf{z}_j) = \exp\!\bigl(-\tfrac12\,\mathbf{z}_i^\top\,\Lambda_{i,j}\,\mathbf{z}_j\bigr)$, yielding the joint distribution:
\begin{equation}
\label{eq:gmrf_natural_form}
p_{\scriptscriptstyle GMRF}(\mathbf{z}) \;\propto\;
\exp\!\Bigl(\,\mathbf{\eta}^\top \mathbf{z}
\;-\;\tfrac12\,\mathbf{z}^\top \Lambda\,\mathbf{z}\Bigr)
\end{equation}
where \(\Lambda = [\Lambda_{i,j}]_{i,j=1}^M\) is the precision matrix and \(\mathbf{\eta} = (\eta_1, \dots, \eta_M)\). Defining \(\mu = \Lambda^{-1}\mathbf{\eta}\) and \(\Sigma = \Lambda^{-1}\) yields the standard form $\mathcal{N}(\mu, \Sigma)$, simplifying both understanding and computation. This Gaussian formulation enables differentiable sampling via the reparameterization trick: using Cholesky factorization \citep{kingma2019introduction} \(\Sigma = L\,L^\top\), we sample as \(\mathbf{z} = \mu + L\,\mathbf{u}\) where \(\mathbf{u} \sim \mathcal{N}(\mathbf{0}, \mathbf{I})\), ensuring fully differentiable backpropagation through the latent space.}

\paragraph{Prior parameterization.}
\textcolor{black}{The prior mean $\boldsymbol{\mu}_p$ and covariance $\boldsymbol{\Sigma}_p$ are learnable parameters. We parameterize $\boldsymbol{\Sigma}_p$ through its lower Cholesky factor $\mathbf{L}_p$, with diagonal entries enforced positive via $\mathrm{diag}(\mathbf{L}_p)=\mathrm{softplus}(\tilde{\mathbf{L}}_p)$ on raw parameters $\tilde{\mathbf{L}}_p$, so that $\boldsymbol{\Sigma}_p=\mathbf{L}_p\mathbf{L}_p^\top$ stays symmetric positive definite during training. Encoder-produced covariances use the same Cholesky reparameterization after SPD projection (Algorithm~\ref{alg:dsp_cov_algo}). Further implementation notes appear in Appendix~\ref{app:technical_experiments}.}

\subsubsection{Conditional Generation}
The GMRF assumption also simplifies conditional sampling. In a
multivariate Gaussian distribution, conditioning on any subset of variables yields another
Gaussian in closed form. Specifically, let
\(\mathbf{z}=(z_1,\dots,z_n)\sim\mathcal{N}(\boldsymbol{\mu},\boldsymbol{\Sigma})\),
where \(\boldsymbol{\mu}=(\mu_1,\dots,\mu_n)\) and
\(\boldsymbol{\Sigma}\) is partitioned into blocks \(\Sigma_{i,j}\). Then,
for indices \(i \neq j\),
\begin{equation}
\label{equ:cond_gen}
p(\mathbf{z}_i \,\big\vert\, \mathbf{z}_j=z_j)
\;=\;\mathcal{N}\bigl(\hat{\mu}_i,\;\hat{\Sigma}_{i,i}\bigr)
\end{equation}
with
\begin{align*}
\hat{\mu}_i &= \mu_i + \Sigma_{i,j}\,\Sigma_{j,j}^{-1}(z_j-\mu_j),\\
\hat{\Sigma}_{i,i} &= \Sigma_{i,i} - \Sigma_{i,j}\,\Sigma_{j,j}^{-1}\,\Sigma_{i,j}^\top
\end{align*}
A full proof is provided in Appendix~\ref{app:conditional_sampling}. In practice, this
closed-form property lets us generate specific components
\(\mathbf{z}_i\) conditioned on partial observations \(\mathbf{z}_j\)
without further training or approximation.

\begin{figure*}[!h]
\centering
\includegraphics[width=.9\textwidth]{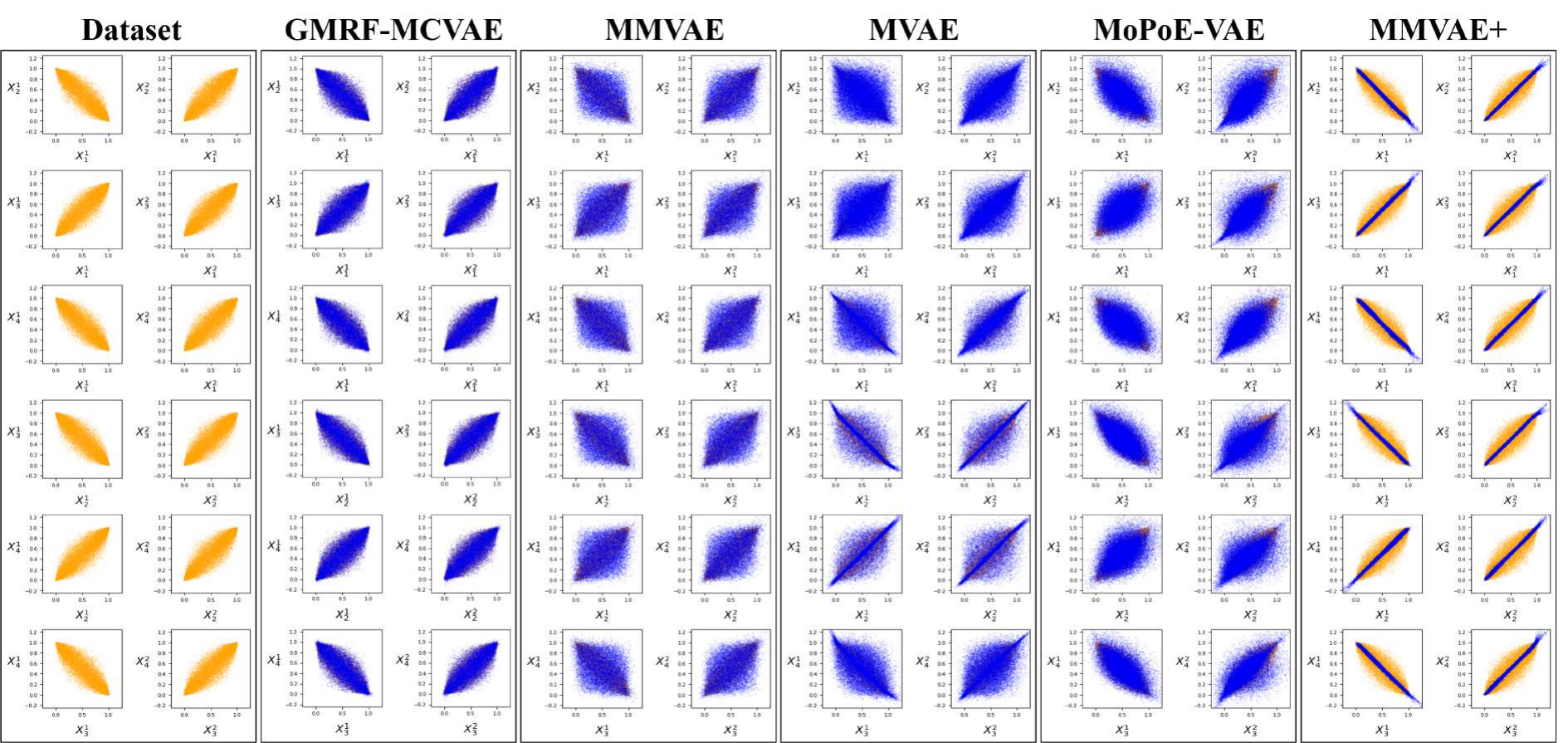} 
\caption{Qualitative results for the unconditional generations on the Copula dataset. Each subplot visualizes joint distributions for each pair of coordinates \((\mathbf{x}_i^1, \mathbf{x}_j^1)\) and \((\mathbf{x}_i^2, \mathbf{x}_j^2)\) across the four two-dimensional components \((\mathbf{x}_1, \mathbf{x}_2, \mathbf{x}_3, \mathbf{x}_4)\). The true distributions are depicted in orange and the generated ones in blue.}
\label{fig:joint_colpula}
\end{figure*}

\subsubsection{Covariance Matrix Construction in the GMRF MCVAE}
\label{sec:covariance_construction}
\textcolor{black}{
To incorporate GMRFs into the MCVAE posterior, we parameterize the latent covariance matrix \(\Sigma = [\Sigma_{i,j}]_{i,j=1}^M\) as a block matrix where \(\Sigma_{i,i} \in \mathbb{R}^{d \times d}\) are component-specific variances from individual encoders, and \(\Sigma_{i,j}\) (\(i \neq j\)) are cross-component covariances from the global encoder. We ensure \(\Sigma\) is symmetric positive definite (SPD) using the following theorem:}

\begin{theorem}
\label{thm:sdp_block}
Consider a block matrix \(\Sigma\) as defined in Section \ref{sec:covariance_construction}. If for each \(i \in \{1, \dots, M\}\) \(\Sigma_{i,i}\) is SPD and satisfies:
\[
||\Sigma_{i,i}^{-1}||^{-1} \geq \sum_{k \neq i} ||\Sigma_{i,k}||
\]
where \(||.||\) is the spectral norm, then \(\Sigma\) is also SPD.
\end{theorem}

\textcolor{black}{
The complete proof is provided in Appendix \ref{app:generalized_hadamard_corollary}. This result establishes the theoretical foundation for our SPD matrix construction method, which we present in Algorithm \ref{alg:dsp_cov_algo}. We defer the algorithmic details and computational complexity analysis to Appendix \ref{app:complexity:baselines}.}

\begin{algorithm}[t]
\caption{SPD Covariance Matrix Construction}
\label{alg:covariance}
\begin{algorithmic}[1]
\Require $\{\Sigma_{i,i}\}_{i=1}^M$ (SPD), $\{\tilde{\Sigma}_{i,j}\}_{i \neq j}$, $\delta > 0$, $\epsilon < 1$
\Ensure SPD matrix $\Sigma$
\For{$i = 1$ to $M$}
    \State $s_i \gets \sum_{j \neq i} ||\tilde{\Sigma}_{i,j}|| + \delta$
    \State $\alpha_i \gets \min(1, \epsilon ||\Sigma_{i,i}^{-1}||^{-1} / s_i)$
\EndFor
\For{all $i,j \in \{1,\ldots,M\}$}
    \State $\Sigma_{i,j} \gets \begin{cases} 
    \Sigma_{i,i} & \text{if } i = j \\
    \tilde{\Sigma}_{i,j} \sqrt{\alpha_i \alpha_j} & \text{otherwise}
    \end{cases}$
\EndFor
\end{algorithmic}
\label{alg:dsp_cov_algo}
\end{algorithm}

\section{EXPERIMENTS}
\label{seq:experiments}

\begin{figure*}[!t]
\centering
\includegraphics[width=1.\textwidth]{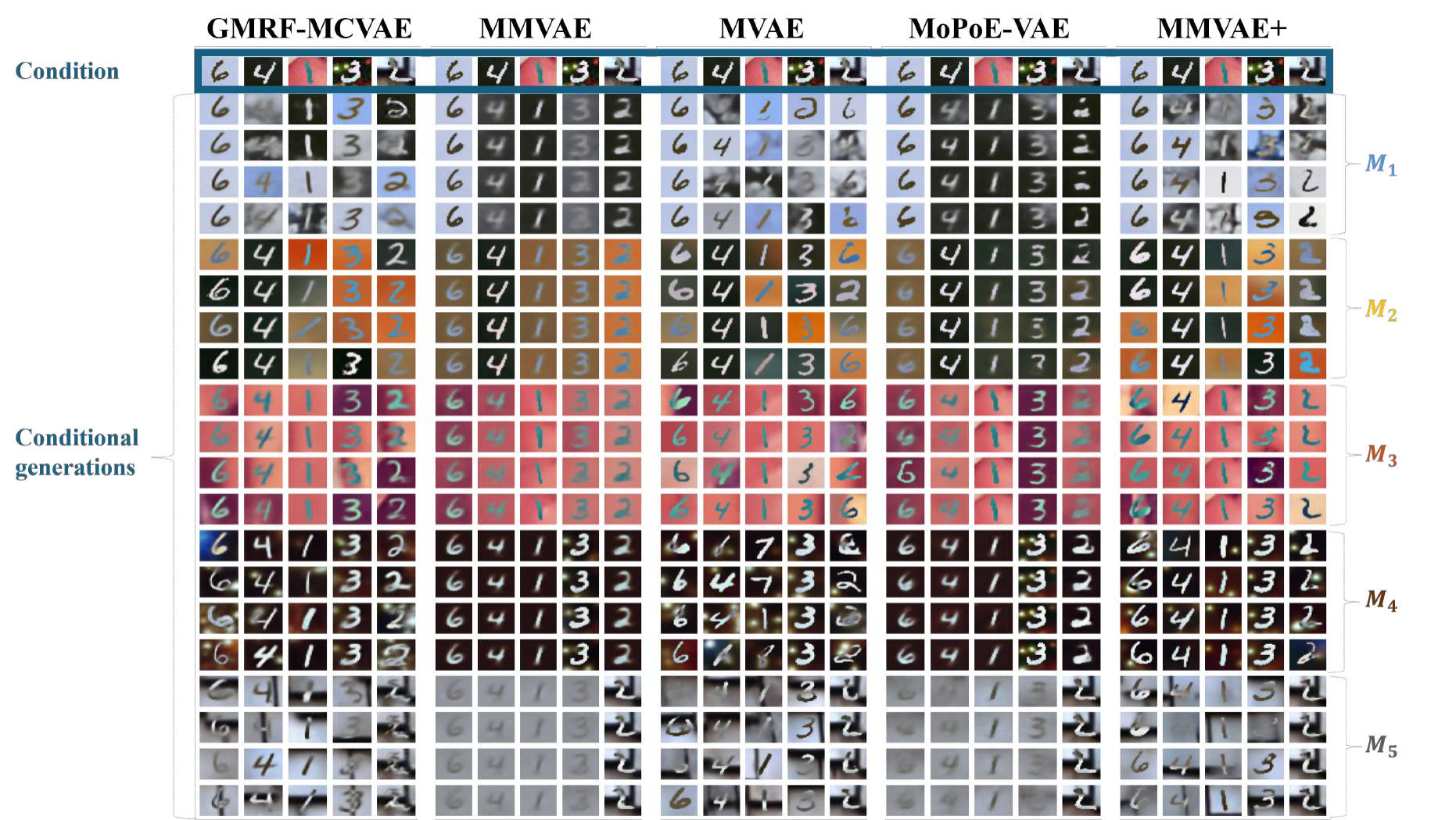} 
\caption{PolyMNIST conditional generations. Each block corresponds to a model. In each column, the first image corresponds to the condition, followed by the conditionally generated components $M_i$.}
\label{fig:conditional_generations_polymnist}
\end{figure*}

We benchmark the proposed GMRF MCVAE against four leading multi-component Variational AutoEncoders: MVAE \citep{wu2018multimodal}, MMVAE \citep{shi2019variational}, MoPoE-VAE \citep{sutter2021generalized}, and MMVAE+ \citep{palumbo2023mmvae}. The core objective is to evaluate the models' ability to learn and preserve complex inter-component structure.

\paragraph{Datasets.} 
\textcolor{black}{We evaluate our GMRF MCVAE on three diverse benchmarks that test different aspects of multi-component modeling. The \textbf{Copula dataset} is a synthetic benchmark we designed using Gaussian Copulas to create components with intricate statistical dependencies while maintaining simple individual distributions \(\mathcal{U}([0,1])\), isolating the challenge of dependency modeling. \textbf{PolyMNIST} \citep{sutter2021generalized} is an established benchmark where five modalities show the same digit with different styles and backgrounds, testing component complexity with simple inter-dependencies. \textbf{BIKED} \citep{regenwetter2022biked} contains real-world bicycle designs with five essential components, requiring models to maintain structural and spatial coherence for industrial feasibility. These datasets span from controlled synthetic dependencies to complex real-world structural constraints.}

\begin{table}[!h]
\centering
\small
\setlength{\tabcolsep}{1mm}
\resizebox{\columnwidth}{!}{%
\begin{tabular}{lcccccc}
\hline
\textbf{Model} & \multicolumn{3}{c}{\textbf{Uncond.\ Gen.}} & \multicolumn{3}{c}{\textbf{Cond.\ Gen.}} \\ \cline{2-7} 
 & \textbf{Dim1} & \textbf{Dim2} & \textbf{Mean} & \textbf{Dim1} & \textbf{Dim2} & \textbf{Mean} \\ \hline
MVAE & \textcolor{black}{2.7{\tiny$\pm$0.4}} & \textcolor{black}{3.2{\tiny$\pm$0.8}} & \textcolor{black}{2.9{\tiny$\pm$0.5}} & \textcolor{black}{3.0{\tiny$\pm$0.3}} & \textcolor{black}{3.1{\tiny$\pm$0.7}} & \textcolor{black}{3.1{\tiny$\pm$0.4}} \\
MMVAE & \textcolor{black}{5.2{\tiny$\pm$0.5}} & \textcolor{black}{4.5{\tiny$\pm$0.4}} & \textcolor{black}{4.8{\tiny$\pm$0.6}} & \textcolor{black}{5.4{\tiny$\pm$0.1}} & \textcolor{black}{4.7{\tiny$\pm$0.3}} & \textcolor{black}{5.0{\tiny$\pm$0.04}} \\
MoPoE & \textcolor{black}{1.9{\tiny$\pm$0.5}} & \textcolor{black}{2.6{\tiny$\pm$0.6}} & \textcolor{black}{2.2{\tiny$\pm$0.4}} & \textcolor{black}{6.0{\tiny$\pm$0.7}} & \textcolor{black}{5.6{\tiny$\pm$0.1}} & \textcolor{black}{5.9{\tiny$\pm$0.3}} \\
MMVAE+ & \textcolor{black}{8.1{\tiny$\pm$0.05}} & \textcolor{black}{4.9{\tiny$\pm$0.07}} & \textcolor{black}{6.5{\tiny$\pm$0.06}} & \textcolor{black}{5.2{\tiny$\pm$0.2}} & \textcolor{black}{4.9{\tiny$\pm$0.1}} & \textcolor{black}{5.1{\tiny$\pm$0.1}} \\
Ours & \textcolor{black}{\textbf{0.7{\tiny$\pm$0.1}}} & \textcolor{black}{\textbf{0.95{\tiny$\pm$0.01}}} & \textcolor{black}{\textbf{0.86{\tiny$\pm$0.04}}} & \textcolor{black}{\textbf{2.6{\tiny$\pm$0.2}}} & \textcolor{black}{\textbf{2.7{\tiny$\pm$0.1}}} & \textcolor{black}{\textbf{2.6{\tiny$\pm$0.1}}} \\
\hline
\end{tabular}%
}
\caption{\textcolor{black}{Mean $\pm$ std} over 3 runs. Wasserstein distance ($\times 10^3$) on the synthetic Copula dataset for unconditional and conditional generation.}
\label{tab:combined_metrics_Copula}
\end{table}

We average all numerical results over three independently trained models. \textcolor{black}{Key tables report mean $\pm$ standard deviation across runs.} Training details, including architectures and hyperparameters, are provided in the Appendix~\ref{app:technical_experiments}. \textcolor{black}{Wall-clock training time per epoch on PolyMNIST is reported in Appendix~\ref{app:wallclock}: despite the covariance construction overhead, our model is faster per epoch than the other baselines.}

\vspace{-1ex}

\begin{figure*}[!t]
\centering
\includegraphics[width=.9\textwidth]{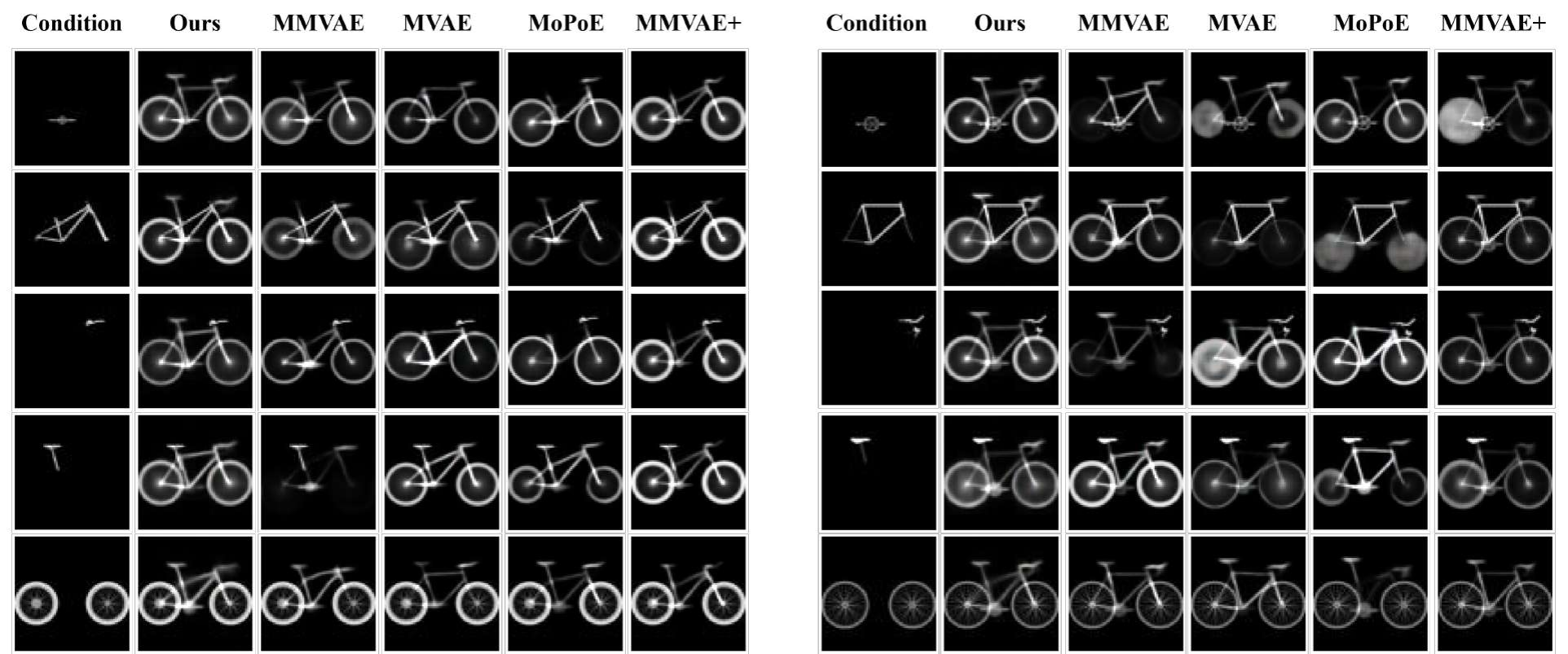} 
\caption{Conditional generation on BIKED. The first column shows the conditioning components, while each subsequent column presents the remaining generated components for each model, overlaid to form the complete bike.}
\label{fig:conditional_generations_biked}
\end{figure*}

\begin{table*}[!t]
\centering
\small
\setlength{\tabcolsep}{4pt}
\begin{tabular}{lcc|ccccc}
\hline
\textbf{Model} & \multicolumn{2}{c|}{\textbf{Unconditional}} & \multicolumn{5}{c}{\textbf{Conditional}} \\ \cline{2-3} \cline{4-8}
 & \textbf{Comp.\,FID\,$\downarrow$} & \textbf{Full\,FID\,$\downarrow$} & \textbf{Comp.\,FID\,$\downarrow$} & \textbf{Full\,FID\,$\downarrow$} & \textbf{Comp.\,SSIM\,$\uparrow$} & \textbf{Full\,SSIM\,$\uparrow$} & \textbf{WS\,$\downarrow$} \\ \hline
MVAE & \textcolor{black}{136.60{\tiny$\pm$1.82}}  & \textcolor{black}{155.06{\tiny$\pm$2.13}}   & \textcolor{black}{134.78{\tiny$\pm$0.81}}  & \textcolor{black}{151.77{\tiny$\pm$3.14}}  & \textcolor{black}{0.85{\tiny$\pm$0.02}} & \textcolor{black}{0.53{\tiny$\pm$0.02}} & \textcolor{black}{0.46{\tiny$\pm$0.03}} \\
MMVAE  & \textcolor{black}{\textbf{133.59{\tiny$\pm$0.78}}}  & \textcolor{black}{\textbf{148.34{\tiny$\pm$1.68}}}   & \textcolor{black}{132.45{\tiny$\pm$0.69}}  & \textcolor{black}{\textbf{146.92{\tiny$\pm$1.05}}}  & \textcolor{black}{0.84{\tiny$\pm$0.01}} & \textcolor{black}{0.50{\tiny$\pm$0.01}} & \textcolor{black}{0.47{\tiny$\pm$0.01}} \\
MoPoE & \textcolor{black}{136.29{\tiny$\pm$3.05}}  & \textcolor{black}{158.25{\tiny$\pm$2.42}}   & \textcolor{black}{131.94{\tiny$\pm$1.77}}  & \textcolor{black}{150.44{\tiny$\pm$1.13}}  & \textcolor{black}{0.85{\tiny$\pm$0.01}} & \textcolor{black}{0.53{\tiny$\pm$0.01}} & \textcolor{black}{0.49{\tiny$\pm$0.02}}\\
MMVAE+  & \textcolor{black}{137.17{\tiny$\pm$1.74}}  & \textcolor{black}{176.12{\tiny$\pm$1.32}}  & \textcolor{black}{131.78{\tiny$\pm$1.21}}  & \textcolor{black}{171.48{\tiny$\pm$2.29}}  & \textcolor{black}{0.84{\tiny$\pm$0.02}} & \textcolor{black}{0.54{\tiny$\pm$0.03}} & \textcolor{black}{0.51{\tiny$\pm$0.09}}\\
Ours & \textcolor{black}{136.83{\tiny$\pm$0.93}}  & \textcolor{black}{186.88{\tiny$\pm$1.40}}   & \textcolor{black}{\textbf{131.52{\tiny$\pm$0.83}}}  & \textcolor{black}{173.70{\tiny$\pm$1.30}}  & \textcolor{black}{\textbf{0.88{\tiny$\pm$0.02}}} & \textcolor{black}{\textbf{0.64{\tiny$\pm$0.02}}} & \textcolor{black}{\textbf{0.31{\tiny$\pm$0.01}}}  \\
\hline
\end{tabular}
\caption{\textcolor{black}{Mean $\pm$ std} (3 runs). FID, SSIM, and center-of-mass Wasserstein (WS) on BIKED. Comp./Full denote component-wise and full-overlap images.}
\label{tab:biked_results}
\end{table*}

\subsection{Copula Dataset Evaluation}
\label{sec:inter-component_coherence}
This subsection evaluates the capability of each model to handle and represent complex inter-component interactions. 

\noindent The synthetic dataset consists of four two-dimensional components, \(\mathbf{x}_1, \mathbf{x}_2, \mathbf{x}_3, \mathbf{x}_4\), each defined as \(\mathbf{x}_i = (\mathbf{x}_i^1, \mathbf{x}_i^2)\) where each component \(\mathbf{x}_i^j\) is uniformly distributed, \(\mathbf{x}_i^j \sim \mathcal{U}([0,1])\), for \(i \in \{1, 2, 3, 4\}\) and \(j \in \{1, 2\}\). The coordinates of each component are generated using two Gaussian Copulas, \(C_j(\mathbf{x}_1^j, \ldots, \mathbf{x}_4^j)\), with uniform means \(\mu_j = [3, \ldots, 3]\) and standard deviations \(\sigma_j = [1, \ldots, 1]\). The correlation matrices \(R^j\) have off-diagonal elements set as \(R_{k,l}^j = ((-1)^j)^{k+l} \cdot 0.9\) \mbox{(Figure \ref{fig:joint_colpula}).}

\vspace{-0.9ex}
\paragraph{Metric} We assess model performance using the Wasserstein distance \citep{villani2009wasserstein}, which measures the optimal transport cost between the empirical probability density functions (PDFs) of the generated and true samples for each component's coordinates. This metric captures differences in both the supports and shapes of distributions. The average of these distances across all comparisons serves as an aggregate performance measure. 

\subsubsection{Qualitative Comparison}
In the analysis of joint distributions (Figure \ref{fig:joint_colpula}), the GMRF MCVAE demonstrates superior alignment with the true distribution, indicating high-quality generations. The MVAE captures certain inter-component relationships more effectively, although this varies across components. Notably, the third component appears less accurate.
This variability is explored in further detail in appendix \ref{appendix:qualitative_results}. As noted by \citet{shi2019variational}, this may stem from the "veto" effect, where experts with higher precision dominate the joint posterior, leading to biased predictions. The MoPoE-VAE combines features of both the MVAE and MMVAE, producing less noisy images but failing to consistently enforce cross-component relationships. Notably, MMVAE+ underperforms compared to other models. While it captures the general correlation between components, it struggles to model the complete dependency structure of the true distribution. This may be due to the increased complexity introduced by component-specific sampling, which complicates the representation of intricate inter-component \mbox{dependencies.}

\vspace{-2ex}

\subsubsection{Quantitative Comparison}
Table \ref{tab:combined_metrics_Copula} confirms the observations from the previous section. The Wasserstein distances between the true and generated distribution PDFs indicate that the GMRF MCVAE generates distributions closer to the true distributions.

\subsection{PolyMNIST Dataset Evaluation}
\label{seq:polymnist_experiment}
In this section, we present the results of our evaluation on the PolyMNIST dataset, assessing the models' ability to handle component complexity and ensure consistency across components. PolyMNIST \citep{sutter2021generalized} extends MNIST into a five-component dataset, where each data point consists of five label-consistent images of the same digit, rendered in distinct handwriting styles and paired with unique background types. This setup challenges models to capture the shared digit identity while managing the complexity introduced by handwriting and background variations. \textcolor{black}{As detailed in Appendix~\ref{app:polymnist_experiment}, we mask $75\%$ of off-diagonal covariance parameters so the number of latent parameters remains comparable to the baselines.}

\paragraph{Metrics}
We evaluate model performance using the Fréchet Inception Distance (FID) \citep{heusel2017gans}, which measures generative quality and diversity \citep{ho2022classifier}, and the Structural Similarity Index (SSIM) \citep{wang2004image}, which assesses perceptual similarity between generated and real images. Additionally, we evaluate global coherence using the methodology proposed by \citet{palumbo2023mmvae}, which measures whether all generated components share the same digit.

\paragraph{Conditional FID protocol.}
\textcolor{black}{Following MMVAE~\citep{shi2019variational} and MMVAE+~\citep{palumbo2023mmvae}, we select one component as observed, encode it, and generate the remaining components. Observed and generated parts are overlaid into a single composite image; FID is then computed between these composites and real test images. All baselines follow the same procedure. SSIM and coherence use the same composite representation.}

\paragraph{Qualitative Comparison}
\label{seq:qualitative_polymnist}
We observe that The GMRF MCVAE model produces consistently complete digits, though slightly blurrier than the MMVAE+ and MVAE (cf Figure \ref{fig:conditional_generations_polymnist}). The MVAE generates sharp images but struggles to align digit identities across components, likely due to the conditional independence assumption induced by the PoE aggregation.
MMVAE however suffers from an averaging effect, which dilutes component-specific details and leads blend backgrounds and a lack of digit variability. 
MoPoE-VAE by construction attempts to balance PoE and MoE, but visually exhibits their combined drawbacks: oversmoothing and incomplete digits.
Despite some trade-offs in sharpness, our GMRF MCVAE consistently preserves the coherence and diversity of the digits, ensuring high generative quality and semantic coherence.

\begin{table}[t]
\centering
\setlength{\tabcolsep}{3pt}
\resizebox{\columnwidth}{!}{%
\begin{tabular}{lccccc}
\hline
\textbf{Model} & \multicolumn{2}{c}{\textbf{Unconditional}} & \multicolumn{3}{c}{\textbf{Conditional}} \\ \cline{2-3}\cline{4-6}
 & \textbf{FID\,$\downarrow$} & \textbf{Coh.\,$\uparrow$} & \textbf{FID\,$\downarrow$} & \textbf{Coh.\,$\uparrow$} & \textbf{SSIM\,$\uparrow$} \\ \hline
MVAE & \textcolor{black}{95.14{\tiny$\pm$0.40}}  & \textcolor{black}{0.139{\tiny$\pm$0.010}}  & \textcolor{black}{94.71{\tiny$\pm$3.20}}  & \textcolor{black}{0.448{\tiny$\pm$0.024}}  & \textcolor{black}{0.993{\tiny$\pm$0.001}} \\
MMVAE  & \textcolor{black}{170.87{\tiny$\pm$3.17}}  & \textcolor{black}{0.175{\tiny$\pm$0.010}} & \textcolor{black}{198.80{\tiny$\pm$2.69}} & \textcolor{black}{0.517{\tiny$\pm$0.010}} & \textcolor{black}{0.995{\tiny$\pm$0.001}} \\
MoPoE-VAE & \textcolor{black}{106.12{\tiny$\pm$1.96}} & \textcolor{black}{0.018{\tiny$\pm$0.001}} & \textcolor{black}{162.74{\tiny$\pm$4.12}} & \textcolor{black}{0.475{\tiny$\pm$0.006}} & \textcolor{black}{0.995{\tiny$\pm$0.001}} \\
MMVAE+  & \textcolor{black}{\textbf{87.23{\tiny$\pm$1.41}}} & \textcolor{black}{0.210{\tiny$\pm$0.006}} & \textcolor{black}{\textbf{82.05{\tiny$\pm$0.78}}} & \textcolor{black}{0.856{\tiny$\pm$0.013}}  & \textcolor{black}{0.994{\tiny$\pm$0.001}} \\
Ours & \textcolor{black}{118.21{\tiny$\pm$1.71}} & \textcolor{black}{\textbf{0.321{\tiny$\pm$0.014}}} & \textcolor{black}{180.76{\tiny$\pm$3.11}} & \textcolor{black}{\textbf{0.869{\tiny$\pm$0.016}}} & \textcolor{black}{\textbf{0.995{\tiny$\pm$0.001}}} \\
\hline
\end{tabular}}
\caption{\textcolor{black}{Mean $\pm$ std} (3 runs). Results on PolyMNIST: FID, coherence (Coh.), and SSIM.}
\label{tab:results}
\end{table}

\paragraph{Quantitative Comparison}
The quantitative evaluation, as presented in Table \ref{tab:results}, reveals that the GMRF MCVAE model performs competitively across all considered metrics on the PolyMNIST dataset. Although the MVAE model achieves the lowest FID scores, the GMRF MCVAE exhibits higher values of cross-coherence and SSIM, suggesting enhanced preservation of structural integrity and global coherence in the generated samples. These results underscore the GMRF MCVAE's ability to produce high-quality, structurally coherent outputs, indicating its robustness in multi-component generative modeling.

\subsection{BIKED Evaluation}
\label{seq:biked}
BIKED \citep{regenwetter2022biked} dataset is a collection of 4,500 bicycle models designed by hundreds of designers, originally introduced for data-driven bicycle design applications. The dataset includes various types of design information, such as parametric data, class labels, and images of full bike assemblies and individual components. In this work, we focus on the segmented component images, where each image corresponds to a different part of a bicycle. Specifically, we retain only the five essential components (saddle, frame, crank, wheels, and handlebars), while discarding components that do not consistently appear across all images, such as water bottles and cargo racks. To reduce GPU load and accelerate training, we convert all images to grayscale and downsample them to 64×64 resolution.

\paragraph{Metrics}
We evaluated generation quality and diversity using FID and structural coherence with SSIM, which measures how well the overall structure of conditionally generated bikes aligns with real ones. While SSIM provides a perceptual evaluation of global shape consistency, industrial design often requires more domain-specific geometric constraints to ensure practical feasibility, e.g., maintaining an appropriate crank-to-wheel distance for maneuverability. To capture such structural dependencies in a scalable and generalizable manner, we propose the Wasserstein distance between component centers of mass, which quantifies the global spatial distribution of parts. \textcolor{black}{Concretely, for each sample we compute the 2D center of mass of every component, stack them into $\mathbf{c}\in\mathbb{R}^{2M}$, and report the 1-Wasserstein distance between empirical distributions of $\mathbf{c}$ for generated and real images.} This approach enables us to assess whether the generated components form structurally plausible assemblies, independent of predefined expert rules, making it applicable to other industrial multi-component datasets.

\paragraph{Conditional FID on BIKED.}
\textcolor{black}{We condition on each of the five components in turn, generate the remaining four, and overlay all five into a full-bike image. FID is averaged over the five conditioning choices. SSIM and the center-of-mass Wasserstein metric use the same overlay representation, with all baselines evaluated identically.}

\subsubsection{Qualitative comparison}
Qualitative results of conditional generation (Figure \ref{fig:conditional_generations_biked}) highlight the limitations of the baseline models. Although MVAE, MMVAE, and MoPoE-VAE exhibit high variability by mixing components from different types of bikes, their generated outputs often lack structural integrity. Specifically, the outputs frequently contain missing or incomplete components, and MVAE, in particular, produces structurally infeasible designs, such as misaligned rear frame sections, causing the rear wheel to be improperly positioned, making the generated bikes impractical for real-world assembly.
MMVAE+ demonstrates better structural consistency, as its generations more closely resemble complete and coherent bicycles when conditioned on specific components. However, this comes at the cost of reduced variability, and we observe frequent missing components, such as absent saddles.
Our model, by contrast, achieves a balance between variability and structural integrity. While its generations may not be as diverse as those of MVAE, they remain structurally coherent and plausible as functional bicycle designs, aligning with expected industrial feasibility.

\subsubsection{Quantitative comparison}
The results in Table \ref{tab:biked_results} align with our qualitative observations. MVAE, MMVAE, and MoPoE-VAE achieve the lowest FID scores, indicating higher diversity in their generations. However, our model maintains competitive FID performance while achieving the best SSIM and Wasserstein distance scores, confirming its ability to generate structurally coherent designs.
Higher SSIM values indicate that our model better preserves local structural consistency in conditional generations, ensuring that the generated components align more closely with real designs. Additionally, the significantly lower Wasserstein distance between the center of mass of the components suggests superior spatial integrity, meaning that the generated parts are correctly positioned relative to each other. This supports our model’s ability to balance diversity with structural plausibility, producing generations that are not only varied but also geometrically coherent and physically plausible.

\textcolor{black}{Block-averaged posterior correlations reveal that mechanically coupled pairs consistently exhibit higher correlation than loosely related ones; see Appendix~\ref{app:biked_correlation} for details.}

\section{CONCLUSION}
This work addressed a core limitation of current multi-component generative models: their inability to explicitly capture dependencies between components. We introduced GMRF MCVAE, integrating Gaussian Markov Random Fields within both prior and posterior distributions to enable structured latent spaces that directly model cross-component relationships.

Our model achieved competitive results on PolyMNIST and outperformed state-of-the-art baselines on both synthetic Copula and real-world BIKED datasets, demonstrating its strength where structural coherence is essential. GMRF MCVAE generates assemblies preserving both diversity and spatial integrity, crucial for industrial design applications.

Future work will investigate sparser MRF structures, improved interpretability, and extensions to diffusion-based generative models.
\subsubsection*{Acknowledgements}
\textcolor{black}{This work was carried out at Centre Borelli, ENS Paris-Saclay, Universit\'e Paris-Saclay, CNRS, in collaboration with Manufacture Fran\c{c}aise des Pneumatiques Michelin. We thank the anonymous reviewers for their constructive feedback.}

\newpage

\bibliographystyle{apalike}
\bibliography{references}

\newpage
\section*{Checklist}



\begin{enumerate}

  \item For all models and algorithms presented, check if you include:
  \begin{enumerate}
    \item A clear description of the mathematical setting, assumptions, algorithm, and/or model. [Yes]
    \item An analysis of the properties and complexity (time, space, sample size) of any algorithm. [Yes]
    \item (Optional) Anonymized source code, with specification of all dependencies, including external libraries. [Yes]
  \end{enumerate}

  \item For any theoretical claim, check if you include:
  \begin{enumerate}
    \item Statements of the full set of assumptions of all theoretical results. [Yes]
    \item Complete proofs of all theoretical results. [Yes]
    \item Clear explanations of any assumptions. [Yes]     
  \end{enumerate}

  \item For all figures and tables that present empirical results, check if you include:
  \begin{enumerate}
    \item The code, data, and instructions needed to reproduce the main experimental results (either in the supplemental material or as a URL). [Yes]
    \item All the training details (e.g., data splits, hyperparameters, how they were chosen). [Yes]
    \item A clear definition of the specific measure or statistics and error bars (e.g., with respect to the random seed after running experiments multiple times). [Yes]
    \item A description of the computing infrastructure used. (e.g., type of GPUs, internal cluster, or cloud provider). \textcolor{black}{[Yes] NVIDIA RTX~3060 Laptop GPU (6\,GB); wall-clock timings in Appendix~\ref{app:wallclock}.}
  \end{enumerate}

  \item If you are using existing assets (e.g., code, data, models) or curating/releasing new assets, check if you include:
  \begin{enumerate}
    \item Citations of the creator If your work uses existing assets. [Yes]
    \item The license information of the assets, if applicable. [No]
    \item New assets either in the supplemental material or as a URL, if applicable. [Not Applicable]
    \item Information about consent from data providers/curators. [Not Applicable]
    \item Discussion of sensible content if applicable, e.g., personally identifiable information or offensive content. [Not Applicable]
  \end{enumerate}

  \item If you used crowdsourcing or conducted research with human subjects, check if you include:
  \begin{enumerate}
    \item The full text of instructions given to participants and screenshots. [Not Applicable]
    \item Descriptions of potential participant risks, with links to Institutional Review Board (IRB) approvals if applicable. [Not Applicable]
    \item The estimated hourly wage paid to participants and the total amount spent on participant compensation. [Not Applicable]
  \end{enumerate}

\end{enumerate}

\clearpage
\appendix
\thispagestyle{empty}
\onecolumn
\section{SUPPLEMENTARY MATERIAL}

\subsection{Conditional Sampling}
\label{app:conditional_sampling}

The conditional distribution of a normally distributed random variable given another is also normally distributed. This is known for the bivariate case in the Matrix Cookbook \citep{petersen2008matrix}. We extend this result for $n \geq 2$, considering a random vector $\mathbf{z} \sim \mathcal{N}(\boldsymbol{\mu}, \boldsymbol{\Sigma})$ with the following probability density function:
\begin{equation}
\mathbf{z} = 
\begin{pmatrix}
\mathbf{z}_1\\
\vdots\\
\mathbf{z}_n
\end{pmatrix}
\sim \mathcal{N}\left(
\begin{pmatrix}
\mu_1\\
\vdots\\
\mu_n
\end{pmatrix},
\begin{bmatrix} 
\Sigma_{11} & \cdots & \Sigma_{1n} \\
\vdots & \ddots & \vdots\\
\Sigma_{n1} & \cdots & \Sigma_{nn} 
\end{bmatrix} 
\right).
\end{equation}
For any pair of indices $i \neq j$ from the set $\{1, \ldots, n\}$, the conditional distribution of $\mathbf{z}_i$ given $z_j$ is
\begin{equation}
p(\mathbf{z}_i|\mathbf{z}_j= z_j) = \mathcal{N}(\hat{\mu}_i, \hat{\Sigma}_{i,i}),
\end{equation}
where
\begin{equation}
\left\{
\begin{aligned}
\hat{\mu}_i &= \mu_i + \Sigma_{i,j}\Sigma_{j,j}^{-1}(z_j - \mu_j), \\
\hat{\Sigma}_{i,i} &= \Sigma_{i,i} - \Sigma_{i,j}\Sigma_{j,j}^{-1}\Sigma_{j,i}.
\end{aligned}
\right.
\end{equation}



\begin{proof}
Consider a random vector $\mathbf{z}$ and distinct indices $i$ and $j$. Define the transformation $\mathbf{y} = A \mathbf{z}_i + B \mathbf{z}_j$ such that $\mathbf{y}$ and $\mathbf{z}_j$ are independent. To achieve $\mathrm{cov}(\mathbf{y}, \mathbf{z}_i) = 0$, it follows that
\[
A \Sigma_{i,j} + B \Sigma_{j,j} = 0.
\]
Selecting $A = I$, leads to
\[
B = -\Sigma_{i,j} \Sigma_{j,j}^{-1}.
\]
Substituting back, we obtain
\[
\mathbf{y} = \mathbf{z}_i - \Sigma_{i,j} \Sigma_{j,j}^{-1} \mathbf{z}_j.
\]

The independence implies $\mathbf{E}[\mathbf{y}|\mathbf{z}_j] = \mathbf{E}[\mathbf{y}] = \mu_i$. Consequently, the conditional expectation of $\mathbf{z}_i$ given $z_j$ is
\begin{align*}
\mathbf{E}[\mathbf{z}_i|z_j] &= \mathbf{E}[\mathbf{y} + \Sigma_{i,j} \Sigma_{j,j}^{-1} \mathbf{z}_j|z_j] \\
&= \mathbf{E}[\mathbf{y}|z_j] + \Sigma_{i,j} \Sigma_{j,j}^{-1} z_j \\
&= \mu_i + A (\mu_j - z_j).
\end{align*}

For the variance, we derive:
\begin{align*}
\mathrm{var}(\mathbf{z}_i|z_j) &= \mathrm{var}(\mathbf{y} - B\mathbf{z}_j|z_j) \\
&= \mathrm{var}(\mathbf{y}|z_j) + \mathrm{var}(B\mathbf{z}_j|z_j) \\
&\quad - B\mathrm{cov}(\mathbf{y}, -\mathbf{z}_j) - \mathrm{cov}(\mathbf{y}, -\mathbf{z}_j)B' \\
&= \mathrm{var}(\mathbf{y}|z_j) \\
&= \mathrm{var}(\mathbf{y}).
\end{align*}

Thus:
 
\begin{align*}
\mathrm{var}(\mathbf{z}_i|z_j) &= \mathrm{var}(\mathbf{z}_i + B\mathbf{z}_j) \nonumber\\
&= \mathrm{var}(\mathbf{z}_i) + B\mathrm{var}(\mathbf{z}_j)B' + B\mathrm{cov}(\mathbf{z}_j, \mathbf{z}_i) \nonumber\\
&\quad - \mathrm{cov}(\mathbf{z}_i, \mathbf{z}_j)B' \nonumber\\
&= \Sigma_{i,i} + B\Sigma_{j,j}B' - B\Sigma_{j,i} - \Sigma_{i,j}B' \nonumber\\
&= \Sigma_{i,i} - \Sigma_{i,j}\Sigma_{j,j}^{-1}\Sigma_{j,i}
\end{align*}

This final expression for $\mathrm{var}(\mathbf{z}_i|\mathbf{z}_j = z_j)$ is the variance of the conditional distribution $p(\mathbf{z}_i|\mathbf{z}_j = z_j) = \mathcal{N}(\hat{\mu}_i, \hat{\Sigma}_{i,i})$, where $\hat{\Sigma}_{i,i} = \Sigma_{i,i} - \Sigma_{i,j} \Sigma_{j,j}^{-1} \Sigma_{j,i}$.
\end{proof}

\subsection{Demonstration of Theorem 1}
\label{app:generalized_hadamard_corollary}
\subsubsection{Preliminaries}

This section demonstrates how our architecture can generate full diagonal block covariance matrices while \mbox{ensuring} symmetric positive definiteness.

\noindent Let \( M \) be a complex matrix, and let \( E \) and \( F \) be two vector spaces equipped with norms \( \|\cdot\|_E \) and \( \|\cdot\|_F \), respectively, such that for all \( \mathbf{x} \in E \), \( M\mathbf{x} \in F \).

The standard operator norm of \( M \) is defined as:
\[
\|M\| = \sup_{\mathbf{x} \neq 0} \frac{\|M\mathbf{x}\|_F}{\|\mathbf{x}\|_E}.
\]

\noindent Consider a block matrix \(\Sigma\) defined as:
\begin{equation}
\label{equ:block_dominant_}
    \Sigma = \begin{bmatrix} 
        \Sigma_{1,1} & \cdots & \Sigma_{1,n} \\
        \vdots & \ddots & \vdots \\
        \Sigma_{n,1} & \cdots & \Sigma_{n,n} 
    \end{bmatrix},
\end{equation}

\noindent where \( \Sigma_{i,i} \) are square matrices and \( \Sigma_{i,j} \) for \( i \neq j \) are rectangular matrices, all with complex entries. 
\noindent In the remainder of this section, we will refer to \( \Sigma \) as defined in Equation~\ref{equ:block_dominant_}.

\noindent Following \citet{feingold1962block}, we say that \(\Sigma\) is block diagonally dominant with respect to the norm \(||\cdot||\) if:
\begin{equation}
\label{equ:block_dominance}
    ||\Sigma_{i,i}^{-1}||^{-1} \geq \sum_{\substack{k=1 \\ k \neq i}}^n ||\Sigma_{i,k}||, \quad \forall i \in \{1, \dots, n\}.
\end{equation}

\noindent In order to proove Theorem \ref{thm:sdp_block}, we need to present first the following two key theorems from \citep{feingold1962block}:

\begin{theorem}
\label{thm0}
A block matrix \(\Sigma\) is nonsingular if it is block strictly diagonally dominant (i.e., strict inequality holds in Equation~\ref{equ:block_dominance}).
\end{theorem}

\begin{theorem}
\label{thm00}
For a block matrix \(\Sigma\), each eigenvalue \(\lambda\) satisfies:
\[
\left(||(\Sigma_{i,i} - \lambda \mathbf{I})^{-1}||\right)^{-1} \leq \sum_{\substack{k=1 \\ k \neq i}}^n ||\Sigma_{i,k}||,
\]
for at least one \(i \in \{1, \dots, n\}\).
\end{theorem}

\subsubsection{Proof of Theorem \ref{thm:sdp_block}}
\textcolor{black}{We outline the logic before the detailed steps: block strict diagonal dominance (Theorem~\ref{thm0}) gives nonsingularity; Theorem~\ref{thm00} locates each eigenvalue $\lambda$ near some diagonal block spectrum; combining the spectral-norm bounds shows every eigenvalue must be positive because each diagonal block is SPD. The matrix is therefore SPD.}
\begin{proof}

To prove Theorem~\ref{thm:sdp_block}, we show that \(\Sigma\) is positive definite.

1. \textit{Nonsingularity}: From Theorem~\ref{thm0}, the block strictly diagonal dominance of \(\Sigma\) guarantees that it is nonsingular.

2. \textit{Positivity}: Let \(\lambda\) be an eigenvalue of \(\Sigma\). Theorem~\ref{thm00} ensures that there exists at least one \(i \in \{1, \dots, n\}\) such that:
\[
\left(||(\Sigma_{i,i} - \lambda \mathbf{I})^{-1}||\right)^{-1} \leq \sum_{\substack{k=1 \\ k \neq i}}^n ||\Sigma_{i,k}||.
\]
Since we consider in our theorem that \(||\cdot||\) is the spectral norm, we have:
\[
||(\Sigma_{i,i} - \lambda \mathbf{I})^{-1}|| = \sup_{j \in \{1, \dots, d\}} \left|\frac{1}{\sigma_j^i - \lambda}\right|,
\]
where \((\sigma_j^i)_j\) are the eigenvalues of \(\Sigma_{i,i}\). Let \(k \in \{1, \dots, d\}\) be the index where the supremum is achieved. Substituting this into the inequality gives:
\[
\left(||(\Sigma_{i,i} - \lambda \mathbf{I})^{-1}||\right)^{-1} = |\sigma_k^i - \lambda|.
\]
Using Theorem~\ref{thm0}, we have:
\[
|\sigma_k^i - \lambda| \leq \sum_{\substack{k=1 \\ k \neq i}}^n ||\Sigma_{i,k}|| < ||\Sigma_{i,i}^{-1}||^{-1}.
\]
For the spectral norm:
\[
||\Sigma_{i,i}^{-1}|| = \frac{1}{\sigma_{\text{min}}^i},
\]
where \(\sigma_{\text{min}}^i\) is the smallest eigenvalue of \(\Sigma_{i,i}\). Substituting into the inequality, we get:
\[
|\sigma_k^i - \lambda| < \sigma_{\text{min}}^i.
\]
This implies:
\[
-\sigma_{\text{min}}^i + \sigma_k^i < \lambda < \sigma_{\text{min}}^i + \sigma_k^i.
\]
Since \(-\sigma_{\text{min}}^i + \sigma_k^i \geq 0\), it follows that \(\lambda > 0\), proving that all eigenvalues of \(\Sigma\) are positive. And thus, \(\Sigma \) is SPD, completing the proof.

\end{proof}

\subsection{Technical details for the experiments}
\label{app:technical_experiments}

Throughout all the experiments, we train each model on 3 independent initializations. In this section we provide the experimental details for both PolyMNIST and the Copula experiments.

\subsubsection{PolyMNIST \& BIKED Experiments}
\label{app:polymnist_experiment}

We employ consistent encoder/decoder architectures across all baseline models, using publicly available implementations for MVAE, MMVAE, and MoPoE-VAE from \citep{sutter2020multimodal}, and for MMVAE+ from \citep{palumbo2023mmvae}. We use similar encoders/decoders architectures in both experiments with different parameters reported on table \ref{tab:architecture_parameters}. Our GMRF MCVAE follows a similar ResNet-based design but differs in that we do not employ two separate latent spaces for joint and component specific encodings. Instead, we introduce an additional fully connected network (three layers of 128 ReLU units each, followed by a linear output) to generate the off-diagonal covariance blocks.

To ensure fair comparisons in both the PolyMNIST and BIKED tasks, we configure all factorized baseline models with a 32-dimensional latent space for PolyMNIST and an 8-dimensional latent space for BIKED (both split into shared and modality-specific subspaces). For unfactorized baselines, we use latent spaces of 512 dimensions for PolyMNIST and 16 dimensions for BIKED. In contrast, our GMRF MCVAE architecture employs a fixed latent space dimension of 16 for PolyMNIST and 4 for BIKED. Additionally, in the PolyMNIST experiment, we mask out 75\% of the off-diagonal parameters in the covariance matrix of the GMRF MCVAE, resulting in 660 parameters per distribution, to align with the effective latent capacity of the baselines.

All models are trained for 100 epochs and monitored using coherence and Fréchet Inception Distance (FID). For baseline models, we experiment with \(\beta \in \{5, 2.5,1 \}\) (lower/higher beta values result in poorer performances)
. For the GMRF MCVAE in particular, we explore \(\beta \in \{2.5\times10^{-3},\, 1\times10^{-3},\, 5\times10^{-4},\, 1\times10^{-4}\}\), where \(\beta\) controls the KL term weight in the ELBO \citep{higgins2017beta}. We find that \(\beta=1\times10^{-3}\) yields the best performance in both datasets.

\subsubsection{Copula Dataset Experiment}
\label{app:Copula_experiment}

Table \ref{tab:enc_dec_architecture} presents the architecture details for both the encoders and decoders used in the Copula experiment. To maintain consistency in latent capacities across different models, the GMRF MCVAE was configured with a latent dimension of 2 (yielding a total capacity of 44), while all other models used a latent dimension of 3 (total capacity of 48).
All models were trained for 200 epochs, exploring a range of \(\beta\) values: \(\{2.5, 1, 0.1, 0.05, 0.001\}\). For baseline models, both Gaussian and Laplacian distributions were tested for the prior, posterior, and log-likelihood calculations. Factorized and unfactorized variants were evaluated for MVAE, MMVAE, and MoPoE-VAE.

\begin{table*}[h!]
\centering
\small
\begin{tabular}{|c|c|c|c|}
\hline
\textbf{Component} & \textbf{Layer} & \textbf{Units} & \textbf{Activation} \\
\hline
\multirow{6}{*}{Encoder} & Fully Connected & $2 \times 256$ & ReLU \\
 & Fully Connected & $256 \times 256$ & ReLU  \\
 & Fully Connected - $\sigma_{shared}$ & $256 \times latent$ & Linear\\
 & Fully Connected - $logvar_{shared}$ & $256 \times latent$ & Linear \\
\cline{2-4}
 & (if factorized) Fully Connected - $\sigma_{specific}$ & $256 \times latent$ & Linear  \\
 & (if factorized) Fully Connected - $logvar_{specific}$ & $256 \times latent$ & Linear  \\
\hline
\multirow{3}{*}{Decoder} & Fully Connected & $input \times 256$ & ReLU \\
 & Fully Connected & $256 \times 256$ & ReLU  \\
 & Fully Connected & $256 \times 2$ & Linear \\
\hline
\end{tabular}
\caption{Architecture details of the encoders and decoders used in the Copula experiment.}
\label{tab:enc_dec_architecture}
\end{table*}

\begin{table*}[h]
\centering
\small

\resizebox{\textwidth}{!}{%
\begin{tabular}{lccccc}
\toprule
\textbf{Layer} & \textbf{Input Channels} & \textbf{Output Channels} & \textbf{Kernel Size} & \textbf{Stride} & \textbf{Note}\\
\midrule
Conv2d (conv\_0) & \(fin\) & \(fhidden\) & \(3\times3\) & 1 & Padding=1\\
Activation & -- & -- & -- & -- & LeakyReLU (0.2)\\
Conv2d (conv\_1) & \(fhidden\) & \(fout\) & \(3\times3\) & 1 & Padding=1, bias enabled\\
Activation & -- & -- & -- & -- & LeakyReLU (0.2)\\
\midrule
Shortcut (if \(fin \neq fout\)) & \(fin\) & \(fout\) & \(1\times1\) & 1 & Learned (no padding)\\
\bottomrule
\end{tabular}}
\caption{ResNet Block Architecture. Each block consists of two convolutional layers with LeakyReLU activations and an optional learned shortcut when the input and output channel dimensions differ.}
\label{tab:resnet_block}
\end{table*}

\begin{table*}[h]
\centering
\small

\resizebox{\textwidth}{!}{%
\begin{tabular}{llcl}
\toprule
\textbf{Stage} & \textbf{Layer Type} & \textbf{Output Dimensions} & \textbf{Activation} \\
\midrule
Input & Image & Input size & -- \\
Initial Conv & Conv2d (3 \(\rightarrow\) 32, kernel=3, padding=1) & Input size & -- \\
ResNet Blocks & Sequence of ResNet blocks with AvgPool (3 blocks) & \(nf_0 \times s_0 \times s_0\) & LeakyReLU (0.2) \\
Flatten & View into vector & \(nf_0\cdot s_0^2\) & -- \\
FC Layers mod & Two separate fully-connected layers & Latent Dimension & -- \\
FC Layers joint & Two separate fully-connected layers & Latent Dimension & -- \\
FC Layer off-diag & One fully-connected layer & Latent Dimension & -- \\
\bottomrule
\end{tabular}}
\caption{Encoder Architecture for PolyMNIST and BIKED. The encoder first applies a convolution to the input image, followed by a series of ResNet blocks with average pooling. The final feature map is flattened and processed by fully-connected layers to generate the latent mean and diagonal covariance parameters. For all baseline models except ours, the ``FC Layers mod'' and ``FC Layers joint'' correspond to the modality-specific and shared posterior encodings, respectively. In contrast, our architecture uses an additional fully connected layer ``FC Layer off-diag'' to encode the embedding utilized by the global encoder for generating the off-diagonal elements of the covariance matrix.}
\label{tab:encoder_architecture}
\end{table*}

\begin{table*}[h]
\centering
\small

\label{tab:decoder_architecture}
\resizebox{\textwidth}{!}{%
\begin{tabular}{llcl}
\toprule
\textbf{Stage} & \textbf{Layer Type} & \textbf{Output Dimensions} & \textbf{Activation} \\
\midrule
Input & Latent vector & Latent Dimension (*2 for all models but ours) & -- \\
FC Layer & Fully-connected & \(nf_0 \times s_0 \times s_0\) & -- \\
Reshape & Reshape to tensor & \(256\times8\times8\) & -- \\
Upsampling & Sequential ResNet blocks with Upsample (scale factor=2) & Gradually upsample to \(nf \times 64 \times 64\) & LeakyReLU (0.2) \\
Output Conv & Conv2d (from nf to 3 channels, kernel=3, padding=1) & Output size & -- \\
\bottomrule
\end{tabular}}
\caption{Decoder Architecture for PolyMNIST. The decoder maps a latent vector to a feature map via a fully-connected layer, followed by a sequence of ResNet blocks with upsampling to reconstruct the image.}
\end{table*}

\begin{table}[h]
\centering
\small
\begin{tabular}{lcc}
\toprule
\textbf{Parameter} & \textbf{PolyMNIST} & \textbf{BIKED} \\
\midrule
\multicolumn{3}{l}{\textbf{Encoder}} \\
\(s_0\)            & 7    & 8 \\
\(nf\)             & 64   & 32 \\
\(nf_{\text{max}}\) & 1024 & 512 \\
Image Size         & \(28\times28\) & \(64\times64\) \\
\midrule
\multicolumn{3}{l}{\textbf{Decoder}} \\
\(s_0\)            & 7    & 8 \\
\(nf\)             & 64   & 32 \\
\(nf_{\text{max}}\) & 512  & 256 \\
Output Size        & \(28\times28\) & \(64\times64\) \\
\bottomrule
\end{tabular}
\caption{Architecture parameters for GMRF MCVAE. Our model uses a single latent space with an extra FC layer for off-diagonal embeddings, unlike baselines that require separate joint and modality-specific spaces.}
\label{tab:architecture_parameters}
\end{table}

\subsection{Diagonal bloc construction details}

We remind that the proposed SDP construction algorithm:

\begin{enumerate}
    \item \textbf{Diagonal Blocks}: 
    Each component-specific encoder produces and SPD $\Sigma_{i,i}$. 
    
    \item \textbf{Off-Diagonal Blocks}: 
    The global encoder outputs the off-diagonal blocks \(\tilde{\Sigma}_{i,j} \in \mathbb{R}^{d \times d}\), capturing relationships between components. By symmetry, \(\tilde{\Sigma}_{j,i} = \tilde{\Sigma}_{i,j}^\top\). Stacking these blocks yields:
    \begin{small}
        \begin{equation}
        \label{equ:block_dominant}
            \tilde{\Sigma} = \begin{bmatrix} 
                0 & \cdots & \tilde{\Sigma}_{1,M} \\
                \vdots & \ddots & \vdots \\
                \tilde{\Sigma}_{M,1} & \cdots & 0 
            \end{bmatrix}
        \end{equation}
\end{small}
    
    \item \textbf{Scaling for Positive Definiteness}: 
    To preserve diagonal dominance and ensure positive definiteness, each block row \(i\) is scaled using:
    \begin{small}
            
    \[
        s_i = \sum_{\substack{j=1 \\ j \neq i}}^M ||\tilde{\Sigma}_{i,j}|| + \delta, \quad
        \alpha_i = 
        \min\!\bigl(1, \epsilon\,\tfrac{||\Sigma_{i,i}^{-1}||^{-1}}{s_i}\bigr)
    \]
    Here, \(\delta > 0\) is a small constant to ensure numerical stability, and \(\epsilon < 1\) controls off-diagonal scaling. \textcolor{black}{We use $\epsilon=0.9$ and $\delta=10^{-6}$ in all main experiments; Table~\ref{tab:ablation_delta} reports a small sensitivity grid.} We then compute the scaled off-diagonal blocks as:
    \[
        \Sigma_{i,j} = \tilde{\Sigma}_{i,j} \cdot \sqrt{\alpha_i \alpha_j}
    \]

    \item \textbf{Matrix Assembly}: 
    The final covariance matrix \(\Sigma\) is assembled as:
    \[
        \Sigma = (\Sigma_{i,j})_{i,j}
    \]
    \end{small}
\end{enumerate}

\noindent This construction ensures that \(\Sigma\) remains symmetric and positive definite while effectively modeling components interactions. However, the block-wise method incurs a computational overhead of \(\mathcal{O}(M^{2}d^{3})\). By applying the construction element-wise, the complexity is reduced to \(\mathcal{O}(M^{2}d^{2})\). Section \ref{app:complexity:baselines}  shows that, even with this extra \(\mathcal{O}(M^{2}d^{2})\) term, the overall training cost remains lower than that of other baselines.
Nevertheless, conditional generation using Equation~\ref{equ:cond_gen} requires each component-specific covariance \(\Sigma_{j,j}\) to be self-contained, i.e., independently derivable using only encoder \(j\). The variance construction method disrupts this independence. To address this limitation, we adopt the simplifying assumption that \(\Sigma_{j,j}\) are diagonal matrices.

\subsection{Complexity comparison}
\label{app:complexity:baselines}

Throughout this section we denote by  
\begin{itemize}
  \item $M$: the number of components (encoders/decoders);
  \item $d$: the dimensionality of each component-specific latent;
  \item $C$: the cost (forward \emph{and} backward FLOPs) of a single
        decoder pass on one mini-batch.
\end{itemize}

\paragraph{Per-step complexities.}  
Three baselines considered in our experiments have the following
dominant costs:
\begin{itemize}
  \item \textbf{MMVAE / MMVAE$^{+}$}\,:  
        $M^{2}$ cross-reconstructions  
        $\;\Longrightarrow\;$ $\mathcal{O}(M^{2}\,C)$.
  \item \textbf{MoPoE-VAE}\,:  
        enumeration of $2^{M}$ modality subsets  
        $\;\Longrightarrow\;$ $\mathcal{O}(2^{M}\,C)$.
  \item \textbf{GMRF MCVAE (ours)}\,:  
        one \(\bigl(Md\bigr)^{2}\) covariance build plus
        $M$ self-decodes  
        \[
            \mathcal{O}\!\bigl(M^{2}d^{2} + M\,C\bigr).
        \]
\end{itemize}
After $\Sigma$ is built, all cross-component generations are
obtained by closed-form Gaussian conditioning; no extra decoder passes
are required during training.

\paragraph{When is our method competitive?}
Because MoPoE-VAE has an exponential cost
\(\mathcal{O}(2^{M}C)\), it is already asymptotically dominated by the
polynomial budgets considered here.
We equal the MMVAE/MMVAE$^{+}$ budget when
\begin{equation}
M^{2}d^{2} + M\,C \;<\; M^{2}\,C
\;\;\Longrightarrow\;\;
C \;>\; \frac{M\,d^{2}}{M-1},
\label{eq:threshold}
\end{equation}

\begin{table}[H]
\centering
\small
\begin{tabular}{lccc}
\toprule
Dataset & $M$ & $d$ & Threshold on $C$ \\
\midrule
Copula      & 4 &  2 & $C > 5.3$ \\
PolyMNIST   & 5 & 16 & $C > 320$ \\
BIKED       & 5 &  4 & $C > 20$ \\
\bottomrule
\end{tabular}
\caption{Numerical thresholds obtained from~\eqref{eq:threshold} for the
three experimental settings.  For reference, two fully-connected layers
of size $256\times256$ already cost $\sim2\times10^{5}$ FLOPs, far above
any threshold listed.}
\label{tab:thresholds_app}
\end{table}

Even though GMRF MCVAE incurs an $\mathcal{O}(M^{2}d^{2})$ covariance
step, realistic decoders (convolutional or large FC) comfortably exceed
the thresholds in Table~\ref{tab:thresholds_app}, making our approach more efficient than the 3 baselines while still enabling closed-form partial-to-full
generation.

\subsection{Wall-clock time per epoch (PolyMNIST)}
\label{app:wallclock}

\textcolor{black}{Table~\ref{tab:wallclock} reports average wall-clock time per training epoch on PolyMNIST (batch size 256, three runs, single NVIDIA RTX~3060 Laptop GPU, 6\,GB). Despite the overhead of covariance assembly, GMRF MCVAE is faster per epoch than all baselines in this regime because it avoids cross-modal reconstructions: conditional samples are obtained in closed form from the joint Gaussian latent.}

\begin{table}[ht]
\centering
\small
\begin{tabular}{lc}
\toprule
\textbf{Model} & \textbf{Time / epoch (s)} \\
\midrule
MVAE & \textcolor{black}{$350.6 \pm 3.2$} \\
MMVAE & \textcolor{black}{$306.1 \pm 1.7$} \\
MoPoE-VAE & \textcolor{black}{$313.5 \pm 2.8$} \\
MMVAE+ & \textcolor{black}{$391.5 \pm 3.2$} \\
GMRF MCVAE (ours) & \textcolor{black}{$128.8 \pm 2.1$} \\
\bottomrule
\end{tabular}
\caption{\textcolor{black}{PolyMNIST wall-clock seconds per epoch (mean $\pm$ std, 3 runs).}}
\label{tab:wallclock}
\end{table}

\subsection{Ablations and larger $(M,d)$ timing}
\label{app:ablations}

\textcolor{black}{Table~\ref{tab:ablation_mask} varies the fraction of masked off-diagonal covariance parameters on PolyMNIST. Table~\ref{tab:ablation_diag} compares the full GMRF covariance to a diagonal variant (no cross-component blocks), and Table~\ref{tab:ablation_delta} sweeps $(\epsilon,\delta)$ in Algorithm~\ref{alg:dsp_cov_algo}. Table~\ref{tab:scale_md} reports mean per-epoch time when duplicating modalities to increase $M$ and $d$ (batch size reduced to 32 where needed for memory).}

\begin{table}[ht]
\centering
\small
\begin{tabular}{lcccc}
\toprule
\textbf{Model (\% mask)} & \textbf{FID (Uncond.)} & \textbf{Coh. (Uncond.)} & \textbf{FID (Cond.)} & \textbf{Coh. (Cond.)} \\
\midrule
GMRF (0\%) & \textcolor{black}{$116.43 \pm 1.20$} & \textcolor{black}{$0.290 \pm 0.019$} & \textcolor{black}{$177.71 \pm 0.03$} & \textcolor{black}{$0.853 \pm 0.096$} \\
GMRF (50\%) & \textcolor{black}{$118.25 \pm 2.40$} & \textcolor{black}{$0.310 \pm 0.020$} & \textcolor{black}{$180.23 \pm 1.81$} & \textcolor{black}{$0.861 \pm 0.060$} \\
GMRF (75\%, default) & \textcolor{black}{$118.21 \pm 1.71$} & \textcolor{black}{$0.321 \pm 0.014$} & \textcolor{black}{$180.76 \pm 3.11$} & \textcolor{black}{$0.869 \pm 0.016$} \\
\bottomrule
\end{tabular}
\caption{\textcolor{black}{PolyMNIST masking ablation (mean $\pm$ std).}}
\label{tab:ablation_mask}
\end{table}

\begin{table}[ht]
\centering
\small
\begin{tabular}{lcc}
\toprule
\textbf{Variant} & \textbf{FID (Uncond.)} & \textbf{Coh. (Uncond.)} \\
\midrule
\textcolor{black}{Diagonal $\boldsymbol{\Sigma}$ (no cross blocks)} & \textcolor{black}{$141.00 \pm 0.59$} & \textcolor{black}{$0.12 \pm 0.07$} \\
\bottomrule
\end{tabular}
\caption{\textcolor{black}{Diagonal-covariance ablation on PolyMNIST (conditional Gaussian conditioning is not applicable in the same form).}}
\label{tab:ablation_diag}
\end{table}

\begin{table}[ht]
\centering
\small
\begin{tabular}{lccc}
\toprule
\textcolor{black}{$\delta$ $\backslash$ $\epsilon$} & \textcolor{black}{$1.0$} & \textcolor{black}{$0.9$} & \textcolor{black}{$0.85$} \\
\midrule
\textcolor{black}{$10^{-7}$} & \textcolor{black}{$118.92 \pm 1.72$ / $0.320 \pm 0.020$} & \textcolor{black}{$118.54 \pm 0.96$ / $0.320 \pm 0.015$} & \textcolor{black}{$119.05 \pm 1.83$ / $0.315 \pm 0.047$} \\
\textcolor{black}{$10^{-6}$} & \textcolor{black}{$118.81 \pm 1.70$ / $0.319 \pm 0.089$} & \textcolor{black}{$118.21 \pm 1.71$ / $0.321 \pm 0.014$} & \textcolor{black}{$119.16 \pm 1.11$ / $0.312 \pm 0.016$} \\
\textcolor{black}{$10^{-5}$} & \textcolor{black}{$118.97 \pm 2.18$ / $0.313 \pm 0.048$} & \textcolor{black}{$118.63 \pm 1.35$ / $0.320 \pm 0.013$} & \textcolor{black}{$120.11 \pm 2.02$ / $0.307 \pm 0.058$} \\
\bottomrule
\end{tabular}
\caption{\textcolor{black}{Sensitivity to $(\epsilon,\delta)$ in Algorithm~\ref{alg:dsp_cov_algo} (PolyMNIST: unconditional FID / coherence). Default is $\epsilon=0.9$, $\delta=10^{-6}$.}}
\label{tab:ablation_delta}
\end{table}

\begin{table}[ht]
\centering
\small
\begin{tabular}{ccc}
\toprule
$M$ & $d$ & \textbf{Avg.\ epoch duration (s)} \\
\midrule
\textcolor{black}{10} & \textcolor{black}{32} & \textcolor{black}{$294.0 \pm 18.4$} \\
\textcolor{black}{10} & \textcolor{black}{64} & \textcolor{black}{$408.5 \pm 25.6$} \\
\textcolor{black}{10} & \textcolor{black}{128} & \textcolor{black}{$1899.9 \pm 418.9$} \\
\textcolor{black}{15} & \textcolor{black}{32} & \textcolor{black}{$466.4 \pm 29.2$} \\
\textcolor{black}{15} & \textcolor{black}{64} & \textcolor{black}{$1017.8 \pm 263.7$} \\
\textcolor{black}{15} & \textcolor{black}{128} & \textcolor{black}{$20789.1 \pm 1301.5$} \\
\bottomrule
\end{tabular}
\caption{\textcolor{black}{Synthetic scaling on duplicated PolyMNIST modalities (mean $\pm$ std per epoch; batch 32).}}
\label{tab:scale_md}
\end{table}

\subsection{BIKED block correlations in latent space}
\label{app:biked_correlation}

\textcolor{black}{For each test sample and training run we convert the posterior covariance into a correlation matrix, average over samples and runs, then average entries between latent dimensions belonging to each component pair. We observe consistent relative structure (which pairs are stronger than others) compatible with mechanically coupled parts, while absolute magnitudes stay moderate due to diagonal-dominance regularization ($\epsilon=0.9$ in all main experiments).}

\subsection{Additional Results from the Copula Experiment}
\label{appendix:qualitative_results}

\paragraph{Marginal Distributions}
\begin{figure*}[!t]
\centering
\includegraphics[width=.9\textwidth]{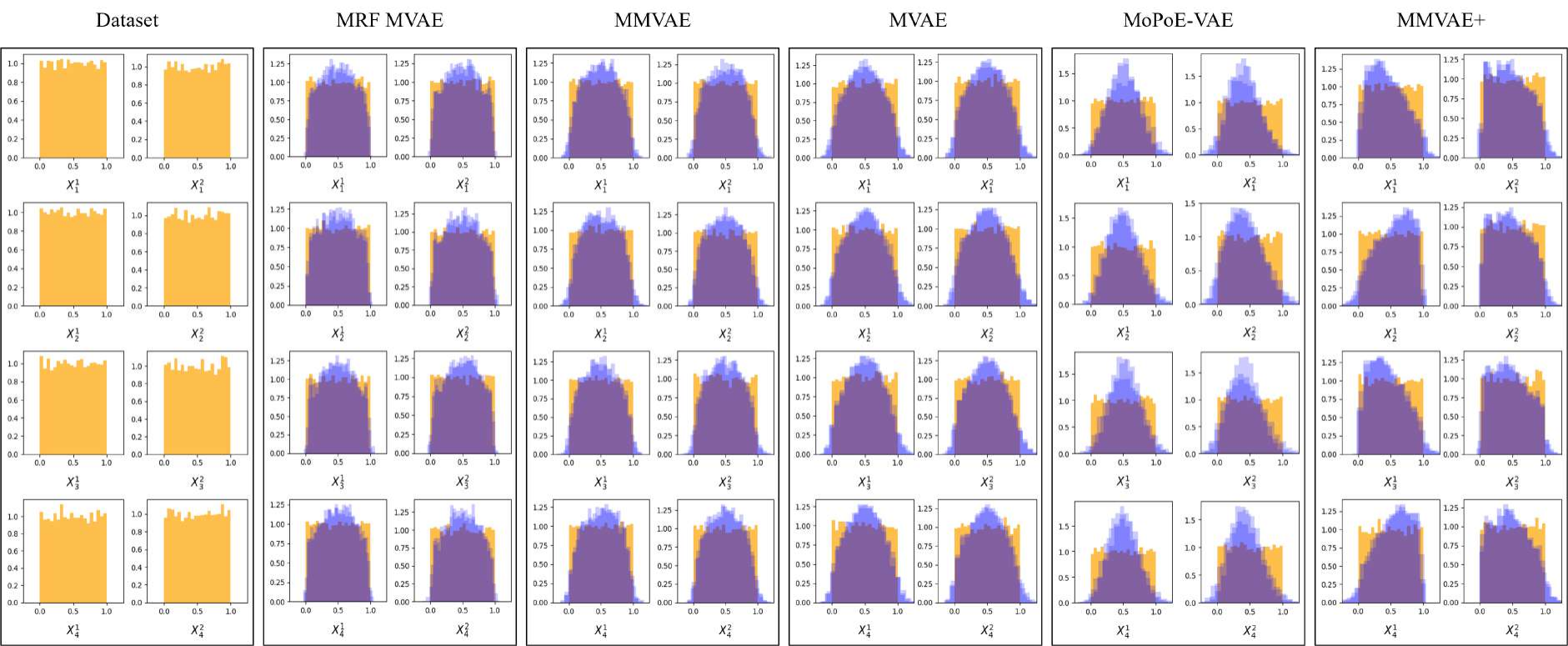} 
\caption{Qualitative analysis of unconditional generations using the Copula dataset. Each subplot displays the marginal distributions for each coordinate: \((\mathbf{x}_i^1)\) on the left and \((\mathbf{x}_i^2)\) on the right, across four two-dimensional components \((\mathbf{x}_1, \mathbf{x}_2, \mathbf{x}_3, \mathbf{x}_4)\). True distributions are depicted in orange and generated distributions in blue.}
\label{fig:marginal_colpula}
\end{figure*}
As shown in Figure \ref{fig:marginal_colpula}, the marginal generations from the various baseline models and the GMRF MCVAE, generally conform to the expected range. Notably, the GMRF MCVAE closely matches the empirical marginal distributions of the dataset, consistently producing outputs within the defined range of [0,1].

\paragraph{Unconditional MVAE Generations}

\begin{figure*}[!t]
\centering
\includegraphics[width=.8\textwidth]{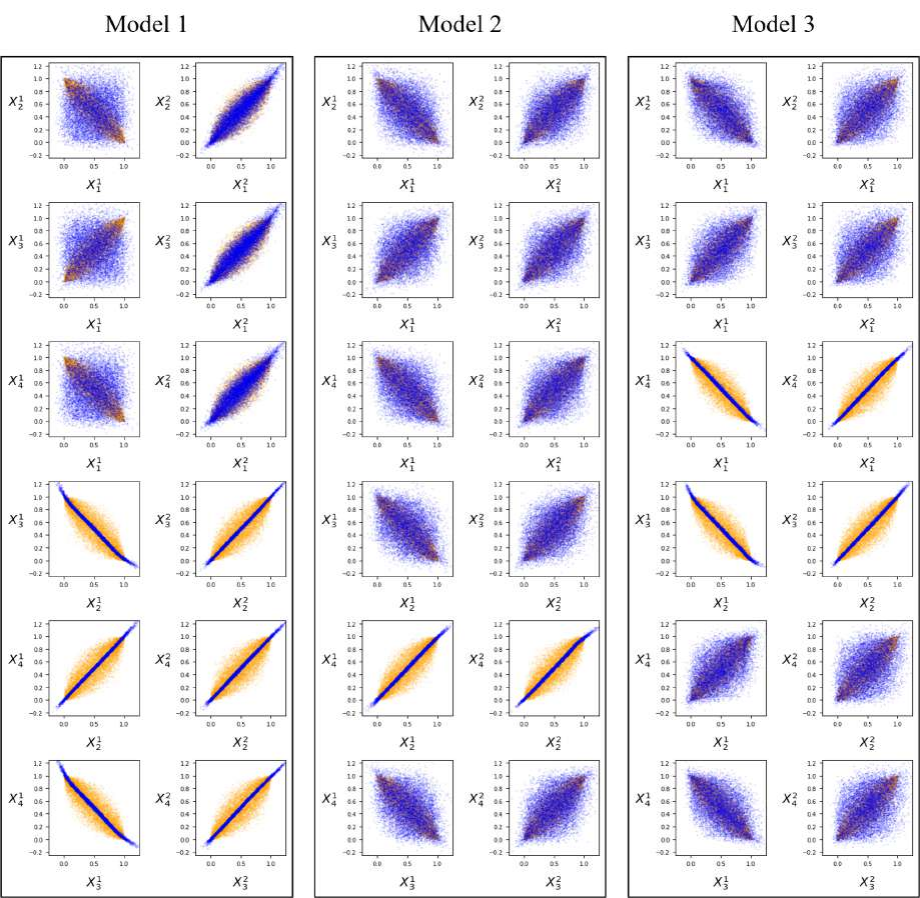} 
\caption{Qualitative results of unconditional generations from the Copula dataset across three training iterations of the MVAE. Each subplot shows joint distributions for pairs of coordinates \((\mathbf{x}_i^1, \mathbf{x}_j^1)\) and \((\mathbf{x}_i^2, \mathbf{x}_j^2)\) across the four two-dimensional components \((\mathbf{x}_1, \mathbf{x}_2, \mathbf{x}_3, \mathbf{x}_4)\). The true distributions are shown in orange, and the MVAE-generated distributions are in blue.}

\label{fig:joint_colpula_mvae}
\end{figure*}

Figure \ref{fig:joint_colpula_mvae} displays the MVAE results after three independent training iterations, revealing inconsistent alignment with the actual joint distributions between components. The MVAE tends to focus selectively on certain components, often overlooking others. This behavior reflects the "veto" effect described in \citet{shi2019variational}, where overconfident experts disproportionately influence the model's output. Such biases negatively impact the global coherence, compromising the accurate representation of inter-component relationships.

\subsection{Exploring a Generalized Variant of GMRF MCVAE Models}

In this section, we discuss a more comprehensive configuration in which both the prior \(p(\mathbf{z})\) and posterior \(p(\mathbf{z}|\mathbf{X})\) are characterized by general Markov Random Fields. This approach opens up possibilities for robustly modeling complex inter-component relationships. However, this configuration also presents significant challenges due to the intractability of the partition functions \(\mathcal{Z}_p\) and \(\mathcal{Z}_q\), which are critical to the prior and posterior distributions. The ELBO for this model configuration is as follows:

\begin{small}
\begin{equation}
    \label{eq:mrf_elbo}
    \begin{split}
    \text{ELBO} &= \mathbb{E}_{q_{\phi}(\mathbf{z}|\mathbf{X})}[p(X|z)] - \log\left(\frac{\mathcal{Z}_p}{\mathcal{Z}_q}\right) \\
    &- \mathbb{E}_{q_{\phi}(\mathbf{z}|\mathbf{X})}\left[\sum_{i<j}\left(\psi_{i,j}^{p}(z_i,z_j) - \psi_{i,j}^{q}(z_i,z_j)\right)\right] \\
    &- \mathbb{E}_{q_{\phi}(\mathbf{z}|\mathbf{X})}\left[\sum_{i}\left(\psi_{i}^{p}(z_i) - \psi_{i}^{q}(z_i)\right)\right]
    \end{split}
\end{equation}
\end{small}

While direct computation of \(\mathcal{Z}_p\) and \(\mathcal{Z}_q\) remains elusive, we can effectively estimate the gradient of the log partition function with respect to the model parameters (\(\theta\)) through sampling. This estimation can be expressed as follows \citep{khoshaman2018gumbolt}:
\begin{small}
\begin{equation}
    \nabla_{\theta} \ln Z_{\theta} = \nabla_{\theta} \ln \sum_{z} \exp \left(-E_{\theta}(\mathbf{z})\right) = -\mathbb{E}_{p_{\theta}(\mathbf{z})}\left[\nabla_{\theta}E_{\theta}(\mathbf{z})\right]
\end{equation}
\end{small}

where \(E_{\theta}(\mathbf{z}) = \sum_{i<j}\psi_{i,j}(z_i,z_j) + \sum_{i}\psi_{i}(z_i)\) represents the energy of configuration \(z\) under the model parameters \(\theta\). This approach enables us to navigate the partition function's intractability, facilitating the model's training through gradient-based optimization techniques.

\end{document}